\documentclass[review]{elsarticle}

\usepackage{hyperref}
\journal{Expert Systems with Applications}

\biboptions{authoryear}
\usepackage{graphicx}
\usepackage{tikz}
\usepackage{hyperref}
\usepackage{rotating}
\usepackage{changes}

\usepackage{subcaption}
\usepackage{multirow}
\usepackage{booktabs}
\usepackage{rotating}

\usepackage{amsmath}
\usepackage{algpseudocode}
\usepackage{algorithm}

\floatname{algorithm}{Procedure}
\algnewcommand\algorithmicforeach{\textbf{for each}}
\algdef{S}[FOR]{ForEach}[1]{\algorithmicforeach\ #1\ \algorithmicdo}

\algdef{SE}[SUBALG]{Indent}{EndIndent}{}{\algorithmicend\ }%
\algtext*{Indent}
\algtext*{EndIndent}

\newcommand{\dstream}{\texttt{D-Stream}}
\newcommand{\dbstream}{\texttt{DBSTREAM}}
\newcommand{\cedas}{\texttt{CEDAS}}
\newcommand{\compass}{\texttt{COMPASS}}

\begin{document}

\begin{frontmatter}

\title{COMPASS: Unsupervised and Online Clustering of Complex Human Activities from Smartphone Sensors}
%\title{COMPASS: Clustering Complex Human Activities from Smartphone Sensors}

\author[affiliation]{Mattia Giovanni Campana\corref{mycorrespondingauthor}}
\cortext[mycorrespondingauthor]{Corresponding author}
\ead{m.campana@iit.cnr.it}

\author[affiliation]{Franca Delmastro}
\ead{f.delmastro@iit.cnr.it}

\address[affiliation]{Institute for Informatics and Telematics of the National Research Council of Italy (IIT-CNR), Via Giuseppe Moruzzi, 1 56124 Pisa, Italy}

\begin{abstract}
Modern mobile devices are able to provide context-aware and personalized services to the users, by leveraging on their sensing capabilities to infer the activity and situation in which a person is currently involved.
Current solutions for context-recognition rely on annotated data and experts' knowledge to predict the user context.
In addition, their prediction ability is strongly limited to the set of situations considered during the model training or definition.
However, in a mobile environment, the user context continuously evolves, and it cannot be merely restricted to a set of predefined classes.
To overcome these limitations, we propose \compass, a novel unsupervised and online clustering algorithm aimed at identifying the user context in mobile environments based on the stream of high-dimensional data generated by smartphone sensors.
\compass can distinguish an arbitrary number of user's contexts from the sensors' data, without defining a priori the collection of expected situations.
This key feature makes it a general-purpose solution to provide context-aware features to mobile devices, supporting a broad set of applications.
Experimental results on 18 synthetic and 2 real-world datasets show that \compass~correctly identifies the user context from the sensors' data stream, and outperforms the state-of-the-art solutions in terms of both clusters configuration and purity.
Eventually, we evaluate its performances in terms of execution time and the results show that \compass~can process 1000 high-dimensional samples in less than 20 seconds, while the reference solutions require about 60 minutes to evaluate the entire dataset.
\end{abstract}

\begin{keyword}
Context-awareness, Unsupervised Machine Learning, Online Clustering, Mobile Computing
\end{keyword}

\end{frontmatter}

% INTRODUCTION -----------------------------------------------------------------------------------------------
\section{Introduction}
\label{section:intro}

Nowadays sensors are pervasive in our daily life since they are embedded in several objects, including houses, cars, and personal mobile devices (e.g., smartphones and wearables).
Moreover, the widespread distribution of personal smart devices and especially the continuous improvement in their sensing capabilities enabled the development of algorithms designed to recognize user activities, thus supporting context-aware applications. Context-awareness refers to the ability of a system to adapt its behavior according to the activity and situation in which the user is currently involved (i.e., her \emph{context}) to improve services' personalization.
These include applications in healthcare~\citep{8674240}, personal assistant development~\citep{8476898}, IoT environments~\citep{8110603}, and recommender systems~\citep{VILLEGAS2018173,CAMPANA201775}.
For example, a mobile device can exploit its sensing capabilities to recognize the situation in which the user is currently involved and, consequently, adapt itself to facilitate the user-device interaction, or to suggest services and contents that match the user's needs in the given context.

According to Peng et al.~\citep{8430551}, human activities can be generally divided into two categories: simple and complex activities.
Simple activities mainly refer to body posture and user's gait (e.g., \emph{walking}, \emph{running}, \emph{sitting}), and they can be characterized by data derived from a small set of sensors embedded in personal and wearable devices like, for example, the accelerometer and the gyroscope.
On the other hand, complex activities have high-level semantics and describe people's daily-life situations, e.g., \emph{working}, \emph{shopping}, or \emph{watching a movie with friends}.
Modeling complex activities requires to combine a heterogeneous set of data, not just including the smartphone physical sensors, but also those data sources that characterize the interactions between the user and her device, and the surrounding environment.
Since in this case, both the terms \emph{context} and \emph{complex activity} refer to the same high-level concept (i.e., the \emph{situation} in which a mobile user is involved), for the sake of simplicity, in the rest of the paper, we use them as synonyms.

The most critical part of a context-aware system is represented by the correct recognition of the user context.
This topic has been widely studied in the literature, and different approaches have been proposed to recognize both simple and, more recently, complex user activities leveraging on mobile sensors data~\citep{6985718, MORALES2017388}.
The great majority of the current solutions are ultimately based on supervised machine learning algorithms that use a data-driven approach to infer the user context.
To this aim, various classification algorithms have been studied, including Decision Trees~\citep{7592098}, Support Vector Machines~\citep{s150820873}, Artificial Neural Networks~\citep{Vaizman:2018:CRI:3178157.3161192}, or ensembles of different classifiers trained on specific context information and then combined with a meta-classifier to infer the general user activity~\citep{8430551}.

Despite their ability to achieve very accurate results~\citep{8090454, MORALES2017388}, context-aware systems based on supervised classifications are strongly limited by the learning process.
In fact, in order to be effective, a supervised model must be trained by using a significant amount of labeled data that represents the different contexts to recognize.
However, both the learning phase and the required labeled data rise three main drawbacks.
First, the collection of (often manually) annotated context data is a time consuming, error-prone, and tedious process that could affect the performance of the learned model~\citep{abdallah2018activity}.
Second, the trained classifier will be able to recognize only the user contexts represented by the labels contained in the training set.
However, the assumption that the training data may represent all the possible situations in which a mobile user could be involved is unrealistic.
Finally, the supervised approach implicitly assumes that the learned model will process sensors data with similar characteristics of the data used during the training phase.
Nonetheless, human activities may vary over time, and dynamic changes in the user context, as well as the emergence of novel situations (not represented in the training dataset), are expected and natural in daily life.

In order to cope with the aforementioned issues, in the last years, researchers have explored different unsupervised approaches to infer the user's high-level context from sensors data.
For example, the use of \emph{ontologies} has been proposed to implement context-aware systems in different pervasive scenarios, including smart home environments~\citep{CIVITARESE201988}, healthcare systems~\citep{Chen2012}, and location-based services~\citep{LEE201741}.
However, even though such a technique does not require a large amount of annotated training data to establish the model, according to~\citep{ye2015usmart}, it mainly relies on the experts' knowledge and usually suffers from the overhead of knowledge engineering.

Another proposed approach to implement unsupervised context-aware systems is based on the use of \emph{clustering} algorithms.
Few researchers have used this approach to identify similarities among sensors data and adapt the system behavior according to the inferred user's context~\citep{LEE201421, 8015566}.
The main drawback of such proposals is that they are ultimately based on traditional (i.e., offline) clustering algorithms, which are not feasible to implement context-aware functionalities in mobile environments.
In fact, traditional clustering assumes that the amount of data to process is fixed, thus the algorithm can evaluate each data sample multiple times to find the clusters configuration that best fits the reference dataset.
However, pervasive systems often depend on sensors that possibly generate thousands of new observations every second.
Therefore, in order to process new observations and possible variations in the data stream, offline clustering algorithms should be periodically re-executed.
This is computationally expensive and requires to store all the sensors data, which is not possible in several scenarios including real-time and mobile applications.

In our research we assume that a context-aware system for mobile and pervasive scenarios must be able to deal with continuous and evolving sensors data to correctly identify the user's context.
Moreover, the user's context must be modeled by using a heterogeneous set of sensors in order to characterize the complex aspects of daily life situations.

To this aim, we propose a novel approach that overcomes the limitations of both supervised and unsupervised context-aware systems proposed so far in the literature.
First of all, it is able to effectively identify the current user's context through the online processing of high-dimensional data streams generated by smartphone sensors.
This allows the mobile system to support a great variety of context-aware applications, since it is not limited to a predefined set of situations.
In addition, the proposed algorithm is characterized by a limited execution time, which
represents one of the most important requirements of context-aware systems for mobile devices.
In fact, due to the characteristics of the mobile environment, the user context may be extremely dynamic.
Therefore, to provide effective responses to the changes in the user situation, the system must be able to identify the user context as soon as possible, by relying on the limited computational resources of the mobile device.

\subsection{Contributions}

\begin{figure}[t]
    \centering
    \resizebox{0.7\columnwidth}{!}{%
        \begin{tikzpicture}
            \node[inner sep=0pt]() at (0,0){
                \includegraphics[width=.9\linewidth]{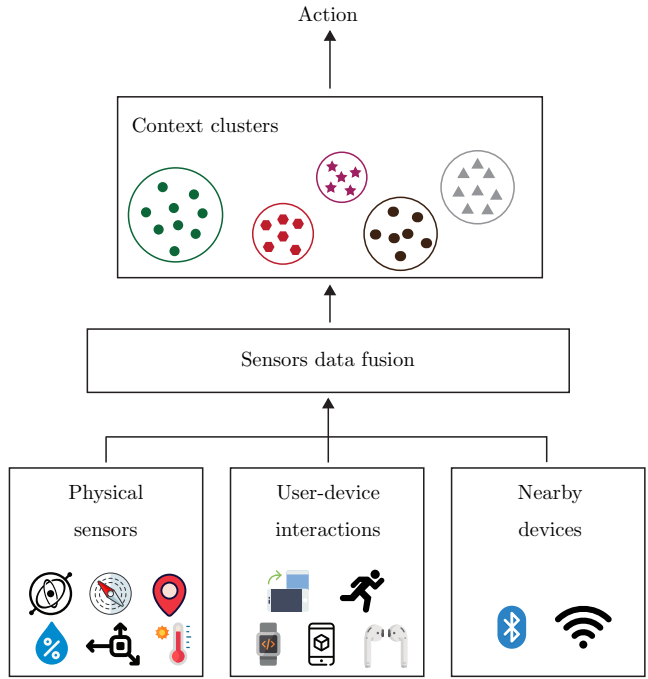}
            };
            \node[] at (0,5.8) {Action};
            \node[] at (-2.1,3.9) {Context clusters};
            \node[] at (0,-0.1) {Sensors data fusion};
            \node[text width=2cm,align=center] at (-3.8,-2.7) {Physical sensors};
            \node[text width=2cm,align=center] at (0,-2.7) {User-device interactions};
            \node[text width=2cm,align=center] at (3.8,-2.7) {Nearby devices};
        \end{tikzpicture}%
    }
    \caption{High-level representation of the context data collection and processing performed by \compass~on the mobile device. The data generated by different mobile sensors are combined and then processed to trigger the most appropriate proactive action according to the identified user context.}
    \label{fig:architecture}
\end{figure}

In this work we present \compass~(Clustering cOMPlex human Activities from Smartphone Sensors), an unsupervised and online algorithm to identify the user's context from mobile sensors data.
Figure~\ref{fig:architecture} shows a graphical representation of the main steps of data collection and processing that induce the system to take the appropriate action.

As a first step, \compass~models the user's context as the combination of heterogeneous features extracted from various data sources, including (i) physical sensors embedded in the mobile device (e.g., GPS, accelerometer, gyroscope), (ii) virtual sensors that describe the interactions between the user and her device (e.g., running applications, and gadgets connected to the smartphone), and (iii) information derived from wireless network interfaces used to identify other devices in proximity.

Then, to adapt its behavior according to the user's current context, the system must  understand whether the feature vector extracted from the sensors represents a situation that has already occurred in the past or not.
To this aim, \compass~implements an online clustering algorithm that processes the data stream generated by the smartphone sensors and extracts groups of similar contexts.
Since mobile devices are memory-constrained, \compass~cannot keep in memory all the data points to identify the relevant clusters.
On the contrary, for each cluster it only keeps track of the most relevant characteristics of the data points that have contributed to its creation, and it updates this information online, while it evaluates new samples coming from the sensors data stream.

\compass~extends the functionalities of \cedas, an online clustering algorithm recently proposed in the literature~\citep{HYDE201796} for non-constrained environments.
\compass~shares with \cedas~the same mechanism to create new clusters online, without processing the data points multiple times.
However, \cedas~requires to set a priori the value of essential parameters, such as the clusters' radius, which depends on the specific characteristics of the final application.
Unfortunately, such knowledge is not available in our reference scenario.
Since the mobile environment is highly dynamic and the user's contexts can follow different and unique characteristics (i.e., data distributions), using the same size for each cluster could strongly limit the algorithm results.
On the contrary, \compass~dynamically adapts its clustering model to the changes in the data stream based on the following assumption: subsequent observations in the data stream might represent the same context and, in this case, they should belong to the same cluster. To this aim, \compass~divides the data stream into time windows adapting the radius of the new clusters based on the characteristics of the observations contained in each specific time window.

In this way, \compass~avoids setting a fixed value for the cluster size by exploiting the time relation and similarity between the context data contained in the stream.
Moreover, it is able to adapt its model to the changes in the user's context by implementing four different mechanisms: (i) finding the best radius size for each cluster, (ii) ``forgetting'' the past and unused clusters, (iii) merging two or more different clusters if they actually represent the same context, and (iv) recognizing new emerging contexts never analyzed before by the algorithm.

Finally, once the current context has been correctly identified, the system can react accordingly.
To this aim, each cluster found by \compass~is associated with one or more proactive \emph{actions}.
These actions can be of different nature:  they can modify the device status (e.g., turning down the ringtone volume during a business meeting), start specific applications (e.g., start a fitness tracker when the user enters into a gym), or even suggest contents and services that can be useful for the user in a specific situation (e.g., a specific playlist to motivate the user during a workout).
Moreover, the actual suitability of the associated actions may depend on the user's current context, and they can be either defined by the mobile user or by third-party applications, or they can be automatically inferred from the actions taken by the user when she has been involved in a similar context in the past.

The association between a context and the set of candidate actions can be easily handled by dynamically creating a set of rules according to the well-known \emph{Event-Condition-Action} (ECA) model~\citep{paton2012active}.
The ECA model has been originally defined in the databases research field, but in the last years, it has been successfully applied in different scenarios, including recommender systems~\citep{LEE201421} and smart home environments~\citep{8574779}.
In this work, we focus on the more challenging problem of modeling the user context from the smartphone sensors, recognizing similar situations that occurred in the past, and identifying new emerging contexts.
By implementing both an online and unsupervised approach, \compass~overcomes the main limitations of standard context-aware solutions based on supervised learning: it does not require neither labeled data to build a model, nor to fix the number of user contexts that the system is able to recognize.
In this regard, \compass~represents a general-purpose solution to provide context-aware features to mobile devices, supporting a broad range of mobile applications.
Moreover, our proposal is lightweight and it can be entirely executed on resource-constrained devices.
Consequently, we can keep the user data on the user's personal device, and the system can provide fast responses according to context's changes.

The main contributions of this work are summarized as follows.

\begin{enumerate}
    \item We propose \compass, an online and unsupervised context recognition algorithm for mobile environments that overcomes the main drawbacks of traditional approaches.
    \item Exploiting the time relation between sensors observations, it presents an advancement of the state of the art in both mobile computing and online clustering research fields.
    \item We compare \compass~performances with three reference online clustering algorithms by using 18 benchmark datasets.
    \item We evaluate \compass~accuracy in the mobile environment by using 2 real-world datasets.
    \item We perform an empirical evaluation of the time complexity of \compass~to evaluate the feasibility of the proposed solution to be entirely executed on a mobile device.
\end{enumerate}

The remainder of this paper is organized as follows.
Section~\ref{sec:related} presents the related work on online clustering, highlighting the main drawbacks that limit the application of state-of-the-art solutions in mobile environments.
Section~\ref{sec:our_solution} describes \compass~in details.
Section~\ref{section:experiment} describes the experimental setting, the metrics and the datasets used to evaluate \compass~performances, and it discusses the experimental results.
Finally, in Section~\ref{sec:conclusions} we draw our conclusions and present some directions for future work.

% RELATED WORK -----------------------------------------------------------------------------------------------
\section{Related Work}
\label{sec:related}

Traditional clustering requires to keep all the data samples in memory and to process them several times to find the best clusters' configuration that optimizes a given metric (e.g., the average squared Euclidean distance in K-means).
In order to implement a context-aware system on mobile devices, a more suitable solution is offered by online clustering algorithms, that process data streams in a more natural way.
In this section we review the main characteristics of the online clustering approach, analyzing the state-of-the-art algorithms and highlighting their main drawbacks concerning their application in our reference scenario.

A data stream is defined as an ordered and potentially unbounded sequence of data points $X = (x_1, x_2, \ldots)$, where $x_t$ is a single observation appeared in the stream at the time $t$.
The underlying distribution of the stream can often change over time and it may exhibit the \emph{concept drift} phenomenon: the position and/or shape of clusters can change, while new clusters may appear and others disappear~\citep{Carnein2019}.
Differently from traditional clustering, online solutions avoid to permanently store all the data points in the stream due to memory constraints.
On the contrary, these algorithms process each data point one time only, updating existing clusters and integrating new observations into the existing model by identifying emerging structures and removing outdated structures incrementally.

Since online clustering represents an effective tool to infer new knowledge from a data stream, the online clustering research field has aroused wide interest in the recent years~\citep{Amini2014, Silva:2013:DSC:2522968.2522981}.
Among the different solutions proposed in the literature over the last years, most data stream clustering algorithms are based on the following two-stage online-offline approach~\citep{Aggarwal:2003:FCE:1315451.1315460}.
First, the algorithm evaluates the incoming data points in real-time and summarizes them by creating a set of \emph{micro-clusters}, which represent groups of similar observations according to a given similarity function.
When a new observation comes from the stream, the corresponding data point is assigned to its closest micro-cluster.
On the contrary, if the data point cannot be assigned to any existing micro-cluster, it will be used to create a new micro-cluster because it can represent the emergence of a new concept in the data stream.
Moreover, during the online phase, the algorithm can also perform additional procedures like merging or deleting micro-clusters to reduce both the computational complexity and memory usage, or to remove noise and outdated information due to the concept drift phenomenon.
Finally, upon request (e.g., the user or application requires a clustering), the offline component of the algorithm determines the final clusters configuration.
This procedure is usually not considered as time-critical, and it is also called \emph{re-clustering phase} because it employs conventional (offline) clustering algorithms to create the \emph{Macro-clusters} by using the micro-cluster centers as data points.

We can divide the online clustering algorithms into two main categories, based on the approach they use to create micro-clusters.
On the one hand, a micro-cluster is represented by its center and some additional information, including its radius and statistics about its data-points (e.g., density and variance).
On the other hand, some algorithms use a grid-based approach to represent micro-clusters.
More specifically, they divide the data space into a grid with equal-sized cells, where a non-empty cell represents a micro-cluster.
~\citep{Carnein:2017:ECS:3075564.3078887} performed an extensive empirical comparison of ten popular online clustering algorithms.
According to their experiments, two algorithms obtained the best results among the others: \dbstream~\citep{7393836} and \dstream~\citep{Chen:2007:DCR:1281192.1281210}.
\dbstream~uses a radius-based approach to create micro-clusters, thus a new observation is assigned to an existing micro-cluster only if it falls within a radius $r$ from its center.
Otherwise, the new observation will be used to initialize a new micro-cluster.
Moreover, to distinguish micro-clusters from simple noise, a cluster must have a density level (e.g., number of points) greater than a specific $C_m$ threshold.
Finally, in the offline phase, \dbstream~merges two micro-clusters into a single Macro-cluster only if they are in close proximity and their overlapping area is greater then a threshold $\alpha$.

\dstream~implements a grid-based approach to find micro-clusters in the data stream.
Grid cells are equally spaced, where a tunable parameter $g_s$ (i.e., \emph{grid size}) specifies the size of each cell.
Moreover, the algorithm estimates the density of each cell by using the number of observations falling in its space.
Based on their density, the grid cells can be classified in the following categories: (i) dense, (ii) transitional, and (iii) sporadic.
These categories can be defined by using two different parameters: $C_m$ and $C_l$, where $0 < C_l < C_m$.
The former density threshold, $C_m$, is used to detect dense grid cells, the latter identifies the sporadic ones, while transitional cells have a density between the two.
The offline procedure of \dstream~ merges adjacent dense cells to form new Macro-clusters and then, the algorithm assigns adjacent transitional cells to their closest Macro-clusters.
Both \dbstream~and \dstream~allows to ``forget'' old and noisy observations.
To this aim, they also depend on a similar parameter called $t_{gap}$ (\emph{gap time}) that specifies how frequently the algorithm should remove micro-clusters with low density and sporadic cells, respectively.

A different approach has been used by \cedas, a fully online clustering solution recently proposed in the literature~\citep{HYDE201796}.
The main goal of the algorithm is to cluster evolving data streams into arbitrarily shaped clusters without relying on the standard online-offline approach.
Similarly to \dbstream, \cedas~assigns a new observation to a cluster only if it falls within a specific radius $r$ from the center of the cluster.
The area between the cluster center and $\frac{r}{2}$ is called \emph{kernel} region, while the rest of the cluster space is defined as \emph{shell} region.
The cluster center is updated to the mean of its samples only when a new observation falls into the shell region.
This limits the micro-cluster movements, thus preventing the single micro-cluster to endlessly following drifting data.
Moreover, based on a density threshold, \cedas~classifies clusters in 2 different categories: (i) micro-cluster, which has a density below the threshold, and (ii) Macro-cluster, with a local density above the threshold.
When a cluster becomes a Macro-cluster, \cedas~checks if its kernel region intersects another cluster shell.
If this is the case, the intersecting clusters are linked to each other in a graph structure and assigned to the same Macro-cluster.
In this way, \cedas~is able to maintain arbitrarily shaped data space regions of Macro-clusters by online processing the data stream, avoiding the traditional online-offline approach.

% PROBLEM STATEMENT ------------------------------------------------------------------------------------------
\section{Problem statement}

\begin{figure*}[t]
    \centering
    \begin{subfigure}{.3\linewidth}
        \centering
        \begin{tikzpicture}
            \node[inner sep=0pt] (russell) at (0,0)
    {\includegraphics[width=\textwidth]{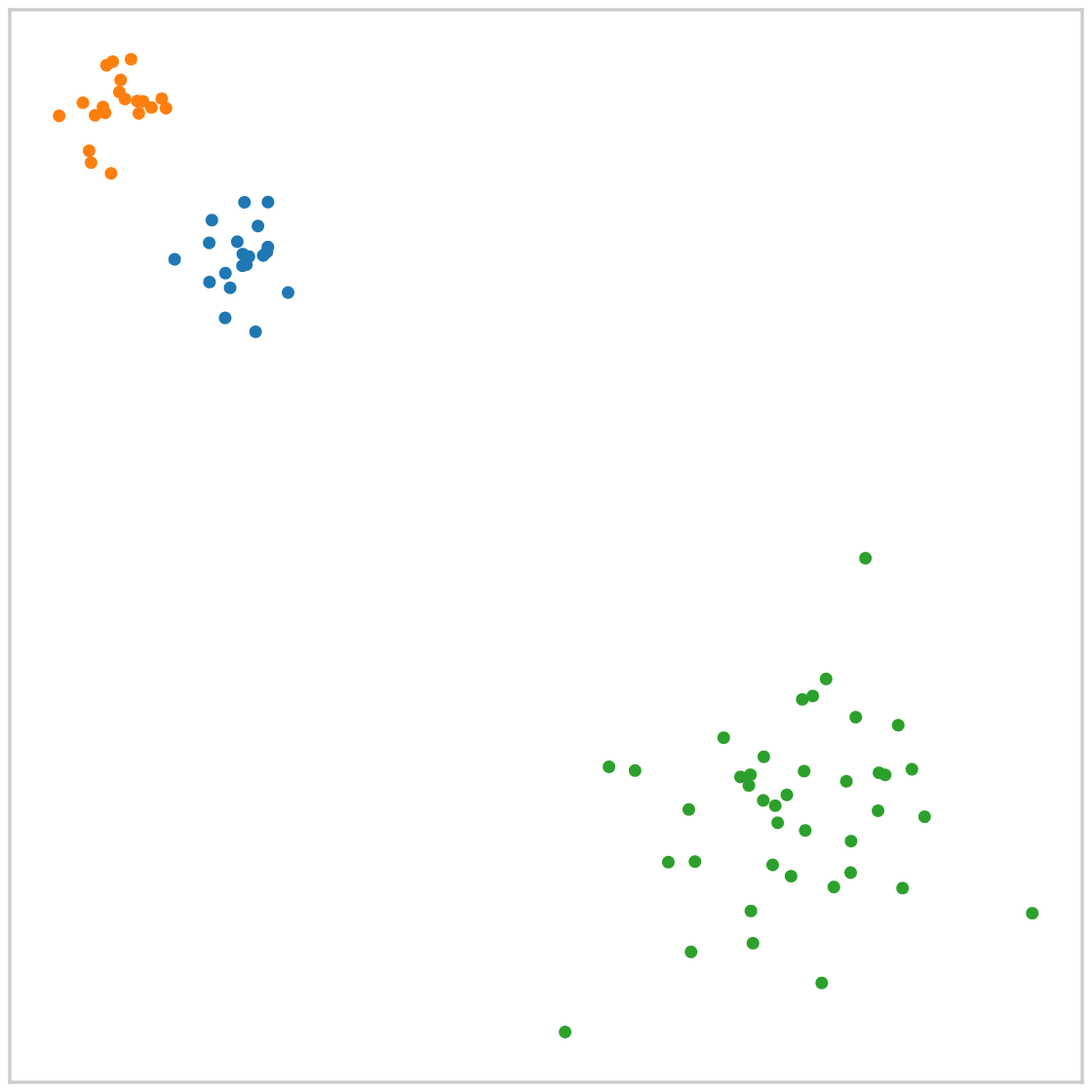}};
        \node[text width=1cm] at (-0.6,1.3) {\texttt{A}};
        \node[text width=1cm] at (-0.3,0.5) {\texttt{B}};
        \node[text width=1cm] at (0.9,-0.2) {\texttt{C}};
        \end{tikzpicture}
        \caption{}
        \label{fig:toy_example}
    \end{subfigure}
    \begin{subfigure}{.3\linewidth}
        \centering
        \includegraphics[width=\linewidth]{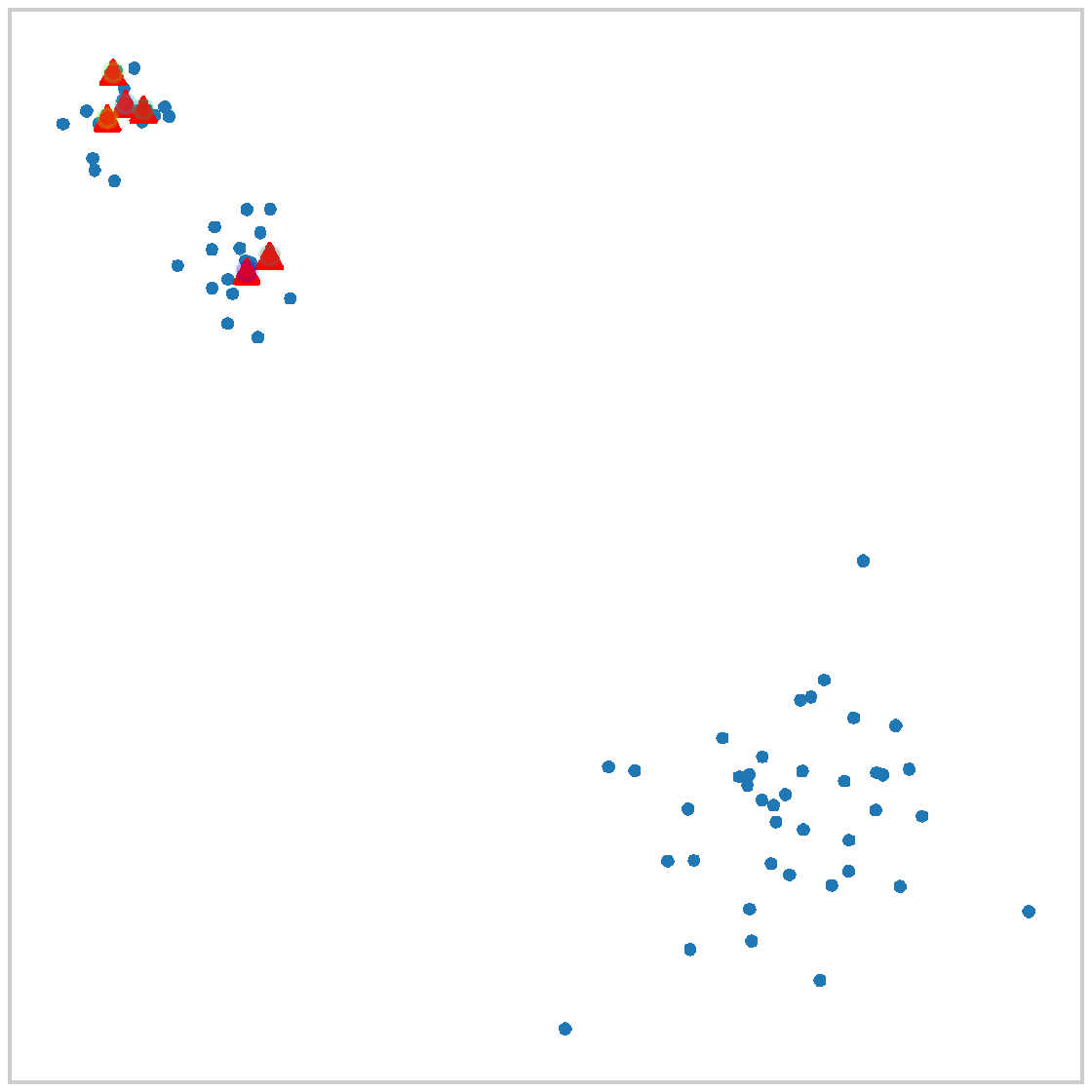}
        \caption{}
        \label{fig:toy_example_cedas1}
    \end{subfigure}
    \begin{subfigure}{.3\linewidth}
        \centering
        \includegraphics[width=\linewidth]{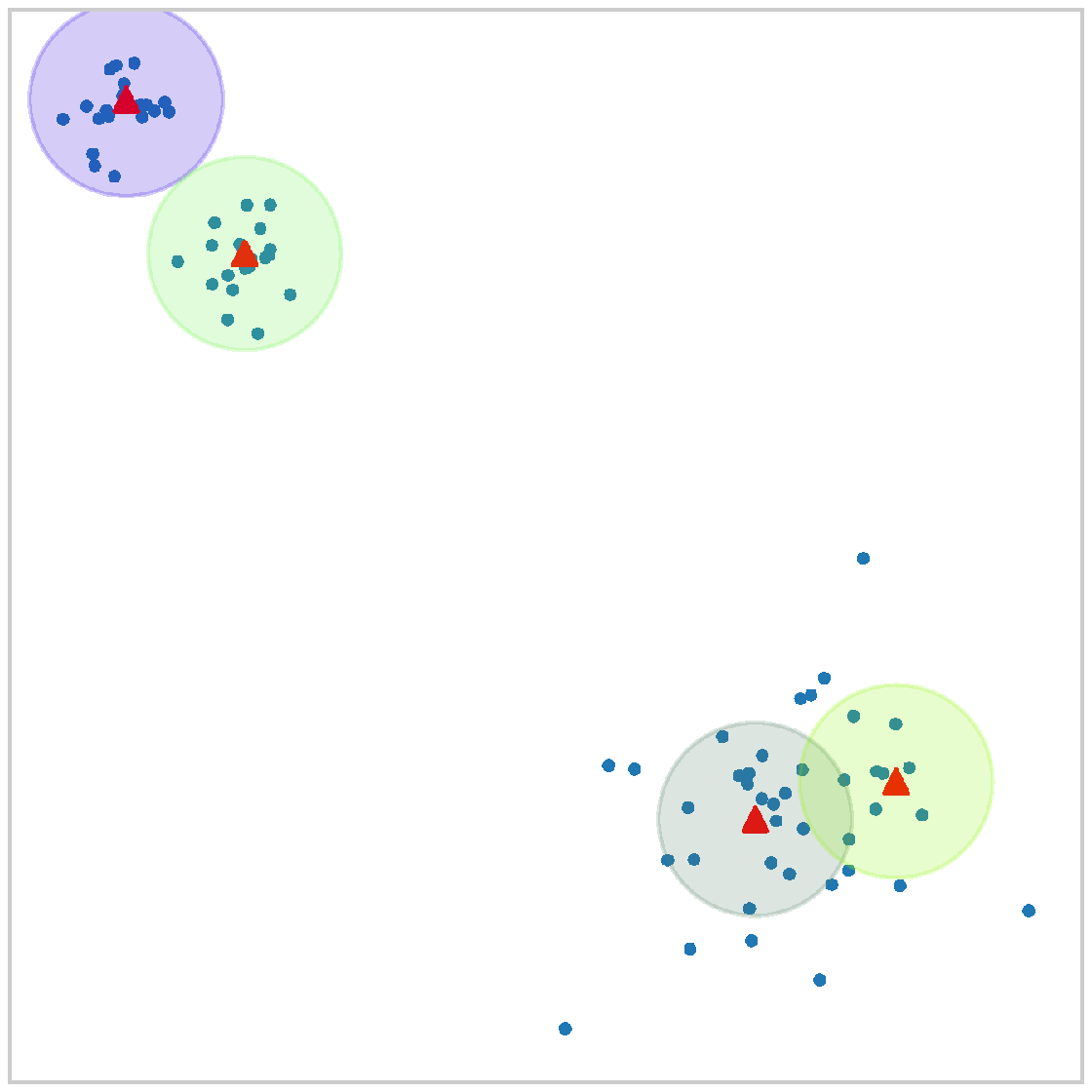}
        \caption{}
        \label{fig:toy_example_cedas2}
    \end{subfigure}\\
    \begin{subfigure}{.33\linewidth}
        \centering
        \includegraphics[width=\linewidth]{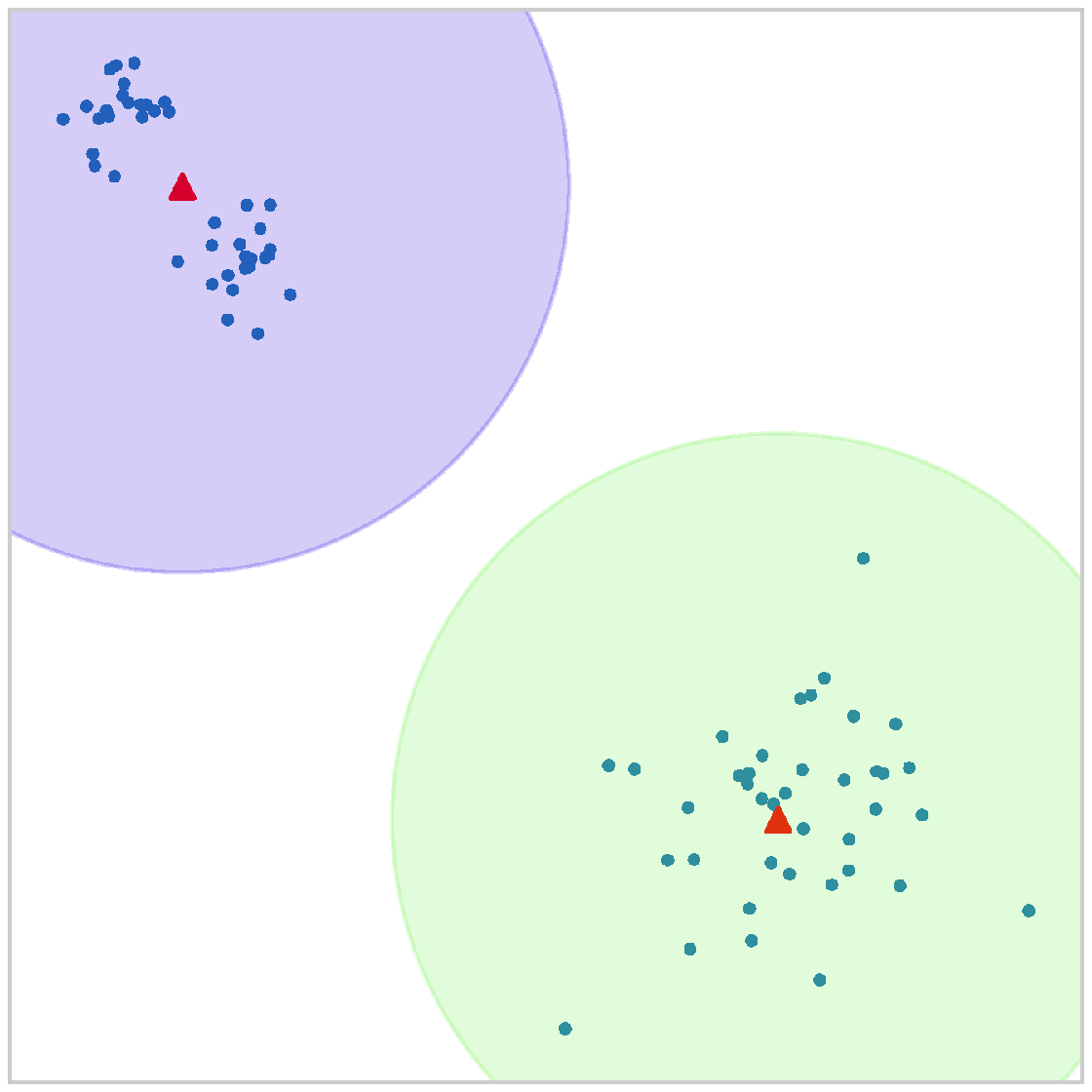}
        \caption{}
        \label{fig:toy_example_cedas3}
    \end{subfigure}
    \begin{subfigure}{.33\linewidth}
        \centering
        \includegraphics[width=\linewidth]{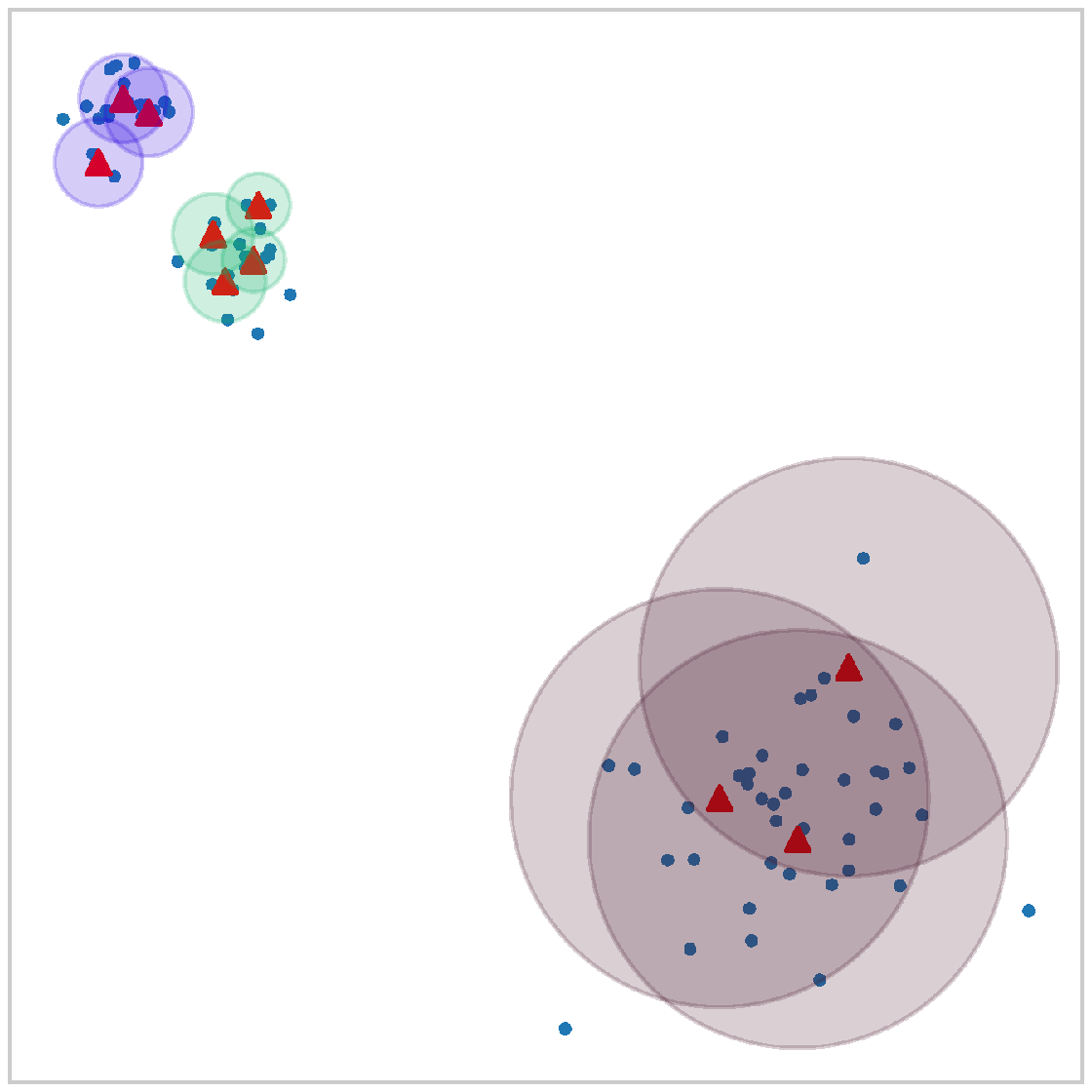}
        \caption{}
        \label{fig:toy_example_our}
    \end{subfigure}
    \caption{Graphical representations of the toy dataset (a), the clusters identified by \cedas~with $r = 0.01$ (b), $r = 0.1$ (c), $r = 0.5$ (d), and those created by \compass~without fixing the radius size (e).
    The red triangles represent the centers of micro-clusters, while the clusters are represented by single circles or overlapping circles with the same color.}
\end{figure*}

One of the main drawbacks of the state-of-the-art online clustering algorithms is the fact that they rely on parameters whose optimal values strongly depend on the specific distribution of data that must be processed.
Specifically, the most important is that used to identify new micro-clusters within the data stream: the parameter $g_s$ for \dstream, and  $r$ for both \dbstream~ and \cedas, which respectively define the size of the grid that partitions the data space and the radius of the new micro-clusters.
Although this parameter is essential to identify the correct clusters' configuration in the data stream, its value is typically static and used to model any new micro-cluster in the stream, regardless of its specific characteristics.
While this approach can be efficient when different clusters follow similar distributions, it represents a limitation when the stream contains groups of data with distinct characteristics that may change over time.

Let us consider, for example, the use of \cedas~to identify the clusters in the toy dataset represented in Figure~\ref{fig:toy_example}.
All the three groups of points belong to the data space defined in $\left[0,1\right]$ and follow a Gaussian distribution, but with different characteristics: the standard deviation of clusters \texttt{A} and \texttt{B} is 0.2, while cluster \texttt{C} is more sparse with a standard deviation of 0.8.
If we set a very small radius (e.g.,  $r=0.01$), \cedas~will identify six different small clusters within the groups \texttt{A} and \texttt{B}, while the majority of the data points will be considered as noise (see Figure~\ref{fig:toy_example_cedas1}).
With a medium-sized radius $r=0.1$ (Figure~\ref{fig:toy_example_cedas2}), the algorithm correctly identifies the clusters \texttt{A} and \texttt{B}, but it splits \texttt{C} in two small and separated clusters.
Finally, with a radius $r=0.5$, it correctly clusters the points of \texttt{C}, but it wrongly combines \texttt{A} and \texttt{B} in one single big cluster (see Figure~\ref{fig:toy_example_cedas3}).

The simple toy example clearly shows that to be efficient, state-of-the-art algorithms must be carefully set up to correctly identify new micro-clusters.
To this aim, they require an exhaustive tuning phase to fit the specific peculiarities of the reference application scenario (e.g., clusters distribution and variability), commonly represented by a training dataset. However, when the reference scenario is utterly dynamic and unpredictable, like the mobile environment, performing the tuning phase does not guarantee the best algorithm configuration.
As we pointed out in the Introduction, assuming that the training dataset may represent all the possible contexts in which a mobile user could be involved is unrealistic. In other words, we cannot expect that the sensors data that represent different user's activities will follow the same distribution nor that their respective clusters in the data stream will have similar characteristics.

Such drawbacks highlight the need for a novel solution that can adapt its behavior according to the changes in the data stream. To this aim, we present \compass~(\emph{Clustering cOMPLex human Activities from Smartphone Sensors}), a novel online clustering algorithm aimed at support clusters' variability by dynamically setting the size of new micro-clusters based on the existing time relation among subsequent observations in the data stream. Compared with the state-of-the-art, COMPASS does not create micro-clusters of the same size but, on the contrary, it continuously adapts them according to the characteristics of the newly processed data.

In the next section, we describe the details of \compass, including its procedures and the method we propose to dynamically change the size of new micro-clusters over time.

% PROPOSED SOLUTION ------------------------------------------------------------------------------------------
\section{COMPASS}
\label{sec:our_solution}

\compass~is a novel online clustering algorithm, especially designed to recognize the user context in mobile environments, based on a broad set of heterogeneous signals generated by different sensors available on modern mobile devices.
It is inspired by \cedas~and shares with it the same process to create Macro-clusters by merging overlapping micro-clusters.
However, instead of using the same radius size for every micro-clusters, \compass~dynamically sets it according to the characteristics of new observations collected from the data stream.

The rationale behind the proposed solution is that context data samples are not independent of each other. On the contrary, if they appear in subsequent time instants, they likely represent the same activity or situation.
Therefore, \compass~aggregates in the same buffer all the data generated by the sensors in a fixed time window.
When the window expires, the algorithm processes all the observations in the buffer to understand if they belong to a previously formed cluster or they represent a new context.
In the former case the algorithm simply updates the previous cluster according to the new observations, while in the latter it must firstly define the dimension of the new micro-cluster.
To this aim, \compass~computes the average distance among the data points in the current buffer as an estimation of the cluster's variability, representing the radius size.
In this way, \compass~is able to continuously adapt its behavior to the changes in the data stream, providing to the mobile device the ability to quickly react to novel contexts, without collecting a big amount of data samples to estimate their distribution.
As a preliminary result, Figure~\ref{fig:toy_example_our} shows the clustering configuration produced by \compass~ with the toy dataset.
Thanks to its ability to dynamically change the micro-cluster size, \compass~finds the best clusters' configuration, in which the formed Macro-clusters exactly match the three groups of points contained in the original dataset.

The full algorithm is composed by four distinct procedures that are sequentially executed for each new data buffer, if required: (i) pre-processing of the time-based buffer, (ii) creation of new micro-clusters, (iii) update of already existing micro-clusters, and (iv) kill old or unused micro-clusters.
In the following sections, we explain the details of each procedure.

\subsection{Buffer pre-processing}

\begin{algorithm}[t]
    \caption{Buffer pre-processing}
    \begin{algorithmic}[1]
        \Require buffer $\mathbf{b}$
        
        \State $r = $ median distance among samples in $\mathbf{b}$
        \ForEach {$s \in \mathbf{b}$}
            \State $c = $ get nearest micro-cluster
            \If{$distance(c, s) \leq r$}
                \State UpdateMicroCluster($c$, $s$)
            \Else
                \State CreateMicroCluster($s, r$)
            \EndIf
        \EndFor
    \end{algorithmic}
    \label{proc: buffer_preprocessing}
\end{algorithm}

Every $t_w$ seconds, \compass~processes all the sensing data collected within the buffer $b$ by running Procedure~\ref{proc: buffer_preprocessing}.
As a first step, the algorithm automatically finds the micro-cluster radius $r$ that best fits the new incoming data samples.
We designed \compass~to process mobile context data that describe the user's daily-life activities. In other words, we assume that the buffer $b$ contains data observations collected from sensors commonly equipped on mobile devices.
Therefore, when $t_w$ is properly set (e.g., 5 or 10 minutes), it is also reasonable to assume that $b$ contains only data samples that refer to the same (or similar) user's context, which implies that they should belong to the same micro-cluster.
For this reason, \compass~calculates $r$ as the mean distance among the sensors' observations contained in $b$.
However, sensor data might be particularly noisy and can negatively affect the creation of the new micro-cluster.
The presence of data points that diverge from the mean distance among those in $b$ (i.e., outliers) may lead to the definition of a too large radius, which can cause the creation of a big micro-cluster that merges two or more distinct micro-clusters representing different user's activities.
Therefore, to reduce the impact of outliers in the radius definition, in Procedure~\ref{proc: buffer_preprocessing} we propose to use the median distance instead of the simple average distance among the data samples contained in $b$.
In this way, only those observations that are closer to each other will actually contribute to the creation of the new micro-cluster, while outliers will be considered as simple noise or the beginning of a new cluster.

Similarly to \cedas, our solution relies on the standard Euclidean distance to divide the data space into several clusters.
However, according to~\citep{7853264}, it is worth noting that the choice of a proper distance metric is crucial in cluster analysis for revealing the natural grouping in a given dataset.
Exploring the use of other distance/similarity metrics and how they can impact on the results produced by \compass~will be part of a future work.

Finally, once the best radius has been set, the algorithm evaluates each data sample $s$ in $b$.
During the execution of this procedure, the algorithm simply checks if the new data point belongs to any current micro-cluster.
If this is the case, \compass~updates the already existing micro-cluster (Procedure~\ref{proc:update_micro-cluster}), otherwise, it executes Procedure~\ref{proc:create_micro-cluster} to create a new data structure that stores all the necessary information to describe the (possibly) new forming cluster.

\subsection{Creation of a new micro-cluster}

\begin{algorithm}[t]
    \caption{\textproc{CreateMicroCluster}$(s, r)$}
    \begin{algorithmic}[1]
        \Require $s, r$
        
        \State $mc = \{$
        
        \Indent
        \State $center = s$
        \State $kernel = [center, \frac{r}{k}]$
        \State $shell = [\frac{r}{k}, r]$
        \State $density = 1$
        \State $edges = list()$
        \State $label = 0$
        \State $time = now()$
        \EndIndent
        
        \State $\}$
        
    \end{algorithmic}
     \label{proc:create_micro-cluster}
\end{algorithm}

As previously explained, \compass~maintains in memory only the information that summarizes the main characteristics of the micro-clusters.
More specifically, for each new micro-cluster, Procedure~\ref{proc:create_micro-cluster} creates a data structure $mc$ that contains the following information:

\begin{description}
    \item[center]: defines the location of the micro-cluster in the data space.
    \item[kernel]: the area of the cluster defined by the data points closest to its center. More specifically, the kernel represents the cluster region between its center and $r/k$, where $k$ is a tunable parameter which governs the actual size of the kernel area with respect to the length of the cluster radius.
    \item[shell]: the cluster area used to find intersections with other micro-clusters in order to create new Macro-clusters; it is defined in the space between $r/k$ and $r$.
    \item[density]: the total number of data samples that lies within the micro-cluster radius and have contributed to its creation.
    \item[edge]: the list of references to the intersecting micro-clusters.
    \item[label]: set to a positive value that uniquely identifies the cluster when the micro-cluster becomes a Macro-cluster.
\end{description}

\subsection{Update micro-clusters}

\begin{algorithm}[t]
    \caption{\textproc{UpdateMicroCluster}$(mc, s)$}
    \begin{algorithmic}[1]
        \Require micro-cluster $mc$, sample $s$
        
        \State $mc.time = now()$
        \State $mc.samples = mc.samples + 1$
        
        \If{$distance(mc.center, s) \leq mc.kernel$}
            \State update $mc.center$
        \EndIf
        
        \If{$mc.samples \geq d$ and $c.label = 0$}
            \State $\mathcal{C} = $ Macro-clusters that intersect with $mc$
            \State $\mathcal{C} = \mathcal{C} \cup \{mc\}$
            \State Create edges among clusters in $\mathcal{C}$
            \State Assign the same label to the clusters in $\mathcal{C}$
        \EndIf
        
    \end{algorithmic}
    \label{proc:update_micro-cluster}
\end{algorithm}

When a new data sample must be assigned to a micro-cluster $mc$, \compass~performs the operations described in Procedure~\ref{proc:update_micro-cluster}.
Specifically, the algorithm firstly checks if the new data point $s$ lies within the micro-cluster kernel.
In this case, the cluster center is updated to the mean of the data samples lying in the kernel region.
Updating the center only when $s$ lies in the kernel region prevents the cluster from endlessly following drifting data by limiting its movement in the data space.

Once the characteristics of $mc$ are updated (i.e., both its center and density values), \compass~checks if it is eligible to become a new Macro-cluster.
When the number of data samples assigned to the cluster is above a fixed $d$ threshold (i.e., density threshold), $mc$ becomes a new Macro-cluster and a new unique label is assigned to it.
Moreover, in this case, the algorithm looks for other intersecting clusters: when a Macro-cluster kernel region intersects the shell region of another Macro-cluster, they are linked together (i.e., the respective edge lists are updated) and they share the same cluster label.
In this way, our algorithm is able to keep track of the changes of arbitrarily shaped clusters in the data stream.

\subsection{Kill micro-clusters}

\begin{algorithm}[t]
    \caption{\textproc{KillMicroClusters}$(t_{max})$}
    \begin{algorithmic}[1]
        \Require $t_{max}$
        
        \ForEach {micro-cluster $mc$}
            \If{$(now - mc.time) < t_{max}$}
                \State Destroy the edges contained in $mc.edges$
                \State Remove $mc$
            \EndIf
        \EndFor
    \end{algorithmic}
    \label{proc:kill_micro-cluster}
\end{algorithm}

As we discussed in Section~\ref{sec:related}, given the dynamic nature of data streams, online clustering algorithms should be able to adapt to concept-drift phenomenons.
This means that the algorithm may update its model not only when new incoming concepts (i.e., clusters) appear, but it should also be able to ``forget'' the outdated ones.

\compass~implements this feature as described in Procedure~\ref{proc:kill_micro-cluster}.
When a micro-cluster $mc$ is not updated for a maximum time threshold $t_{max}$, the algorithm first removes all the edges between $mc$ and other clusters and, eventually, it removes the $mc$ data structure from the model.

% EXPERIMENTS ------------------------------------------------------------------------------------------------
\section{Experiments}
\label{section:experiment}

\begin{table}[t]
\caption{Parameter values explored by using the grid-search approach.}
\centering
\begin{tabular}{ l c l r r}
    Algorithm & Parameter & Values space
    \\\midrule
    \multirow{2}{*}{\dbstream} 
    & $r$       &  $[0, 10]$          \\
    & $C_m$     &  $[1, 10]$          \\
    & $\alpha$  &  $[0, 1]$           \\
    & $t_{gap}$ &  $[100, 5000]$      \\
    \midrule
    \multirow{2}{*}{\dstream}
    & $g_s$     &  $[0, 10]$          \\
    & $t_{gap}$ &  $[100, 5000]$      \\
    & $C_m$     &  $[1, 10]$          \\
    & $C_l$     &  $[0, 10]$          \\
    \midrule
    \multirow{2}{*}{\cedas}
    & $r$       &  $[0.01, 1]$        \\
    & $d$       &  $[2, 20]$          \\
    \midrule
    \multirow{2}{*}{\compass}
    & $t_w$     &  $[60, 1200]$    \\
    & $d$       &  $[2, 20]$          \\
    & $k$       &  $[1, 10]$          \\
    \bottomrule
\end{tabular}
\label{tab:grid_search_space}
\end{table}

In order to evaluate the performance of our proposal, we perform 3 different sets of experiments.
As a first step, we evaluate the effectiveness of \compass~in recognizing clusters of different shapes and densities in a static scenario by using 18 synthetic datasets.
Then, we evaluate the ability of our fully unsupervised and online solution to recognize the user's context by exploiting the sensing capabilities of mobile devices.
In this case we asses the ability of \compass~ to successfully identify evolving clusters in a data stream by using two real-world datasets.
Eventually, we perform an empirical evaluation of the \compass~execution time in order to assess its ability to quickly identify the user context in mobile environments and its suitability of running on resource-constrained devices.

In the experiments we compare the performances of \compass~with the three algorithms that represent the state-of-the-art of online clustering: \dbstream, \dstream, and \cedas.
For \dbstream~and \dstream, we use their implementation in the \texttt{stream} package, a plugin specifically designed to evaluate online clustering solutions by using the statistical programming language \texttt{R}~\citep{stream}, while for \cedas we rely on the \texttt{Matlab} implementation released by the authors~\footnote{\url{https://github.com/RHyde67/CEDAS}}.
Each algorithm relies on a different set of parameters which optimal values may depend on the characteristics of the single dataset.
For this reason, in every experiment we firstly find the best parameter configuration for each algorithm by using the grid-search optimization approach, an exhaustive searching through the parameter spaces. The results are summarized in Table~\ref{tab:grid_search_space}.
Finally, we compare and discuss the performances obtained by the best models.

\subsection{Evaluation metrics}
In our experiments, the true class labels of the data samples are known.
Even though we do not exploit them to define the clustering model, the availability of the ground truth allows us to evaluate the quality of the clustering results by comparing the generated clusters configurations with the groups of data points defined by the labels.
One of the most common evaluation metric used in this case is represented by the Adjusted Random Index (ARI) defined in~\citep{hubert1985comparing} as follows:

\begin{equation}
\small
    ARI = \frac{
\sum_{ij}\binom{n_{ij}}{2} - \left (  \sum_i \binom{a_i}{2} \sum_j \binom{b_j}{2} \right ) / \binom{n}{2}
}
{
\frac{1}{2} \left ( \sum_i \binom{a_i}{2} + \sum_j \binom{b_j}{2} \right ) - \left ( \sum_i \binom{a_i}{2} \sum_j \binom{b_j}{2} \right ) / \binom{n}{2}
},
\label{eq:ari}
\end{equation}

where $a_i$ represents the number of points in a class $i$, $b_j$ denotes the number of data samples in cluster $j$, and $n_{ij}$ is the number of samples in class $i$ and cluster $j$.
The ARI is defined in the range $\left[ -1, 1\right]$, where $0$ indicates that the found clustering configuration is comparable to the results produced by a random guesser, $1$ indicates that the clusters are identical to the groups in the dataset, and $-1$ that the results are worse than by chance.

We use ARI to evaluate the cluster configuration produced by an online clustering algorithm by comparing the set of the generated Macro-clusters with the available ground truth.
Unfortunately, according to~\citep{Carnein:2017:ECS:3075564.3078887}, this metric cannot be used also to assess the quality of the micro-clusters since, usually, the number of created micro-clusters is much larger than the true groups in the data stream.
Therefore, to evaluate the generated micro-clusters, we rely on the Purity metric, defined in~\citep{doi:10.1137/1.9781611972764.29} as follows:

\begin{equation}
    Purity = \frac{
\sum_{i=1}^{K} \frac{\left | C_{i}^{d} \right |}{\left | C_{i} \right |}
}
{K},
\label{eq:purity}
\end{equation}

where $K$ is the number of micro-clusters, $\left | C_{i}^{d} \right |$ denotes the number of points with the dominant class label in micro-cluster $i$, and  $\left | C_{i} \right |$ denotes the total number of points in $i$.

\subsection{Synthetic datasets}

\begin{table}[t]
    \caption{Characteristics of the synthetic datasets.}
    \small
    \begin{tabular}{lrrl}
        \toprule
        Dataset    & Samples & Clusters  & Shapes\\
        \midrule
        square     & 1000        & 4     & spherical                    \\
        spiral     & 1000        & 2     & spirals                      \\
        complex8   & 2551        & 8     & various                      \\
        complex9   & 3031        & 9     & various                      \\
        disk-1k    & 1000        & 2     & concentric spheres                \\
        donut1     & 1000        & 2     & concentric rings and spheres    \\
        donut2     & 1000        & 2     & concentric rings and spheres    \\
        donut3     & 999         & 3     & concentric rings and spheres    \\
        engytime   & 4096        & 2     & spherical and elliptic       \\
        shapes     & 1000        & 4     & various                      \\
        zelnik1    & 299         & 3     & concentric rings and spheres    \\
        zelnik3    & 266         & 3     & various                      \\
        zelnik6    & 238         & 3     & concentric rings and spheres    \\
        r15        & 600         & 15    & spherical                    \\
        rings      & 1000        & 3     & concentric rings                \\
        3mc        & 400         & 3     & various                      \\
        compound   & 399         & 6     & various                      \\
        impossible & 3673        & 8     & squared, spirals, concentric rings, and elliptic \\
        \bottomrule    
    \end{tabular}
    \label{tab:syn_datasets_characteristics}
\end{table}

We start analysing 18 bi-dimensional artificial datasets commonly used for  clustering algorithms benchmarking~\footnote{We selected 18 datasets from a collection of 122 labeled clustering problems that can be downloaded from the following Github source:~\url{https://github.com/deric/clustering-benchmark}}.
Even though the user context in mobile environments is characterized by a high-dimensional features vector, each dataset poses different challenges in terms of recognition of clusters with different shapes and distributions.
For example, Figure~\ref{fig:syn_original} shows the graphical representation of 3 of the selected datasets: \emph{square1}, \emph{spiral}, and \emph{impossible}.
Both \emph{square1}~\citep{handl2004multiobjective} and \emph{spiral}~\citep{handl2004multiobjective} are composed by 1000 data samples, forming respectively four spherical and two spiral-shaped clusters, while the \emph{impossible} dataset is composed by 3673 samples forming 8 clusters characterized by different shapes (i.e., spirals, concentric circles, spherical, and elliptic).

\begin{table}[t]
\caption{Best parameters found for the synthetic datasets. Column $P$ refers to the parameter names, while the numbered columns represent the datasets as follows: $1 = square$, $2 = spiral$, $3 = complex8$, $4 = complex9$, $5 = disk1k$, $6 = donut1$, $7 = donut2$, $8 = donut3$, $9 = engytime$.}
\centering
\small
\begin{tabular}{ l c r r r r r r r r r}
    Algorithm & P & 1 & 2 & 3 & 4 & 5 & 6 & 7 & 8 & 9
    \\\midrule
    \multirow{4}{*}{\dbstream} 
    & $r$ & .66 & .03 & .20 & .88 & .98 & .31 & .86 & .34 & .43\\
    & $C_m$ & 2.27 & 2.87 & 2.91 & 1.43 & 2.51 & 1.52 & 1.18 & 1.36 & 1.55\\
    & $\alpha$ & .30 & .35 & .19 & .35 & .79 & .25 & .55 & .26 & .35\\
    & $t_{gap}$ & 3.1k & 3.9k & 137 & 4.2k & 1.9k & 442 & 4.9k & 850 & 785\\
    \midrule
    \multirow{4}{*}{\dstream}
    & $g_s$ & .27 & .14 & .54 & .31 & .14 & .17 & .30 & .14 & .36\\
    & $t_{gap}$ & 4.9k & 3.1k & 3.6k & 3.6k & 3.8k & 2.1k & 4.8k & 2.8k & 1.1k\\
    & $C_m$ & 1.93 & 1.27 & 1.31 & 2.94 & 1.74 & 2.27 & 1.52 & 2.43 & 2.57\\
    & $C_l$ & .04 & .14 & 1.01 & 1.18 & .78 & 1.34 & .56 & .22 & .58\\
    \midrule
    \multirow{2}{*}{\cedas}
    & $r$ & .1 & .01 & .2 & .1 & .05 & .01 & .01 & .01 & .02\\
    & $d$ & 20 &   2 &  2 & 20 &.  3 &   8 &  8  &   8 &   5\\
    \midrule
    \multirow{3}{*}{\compass}
    & $t_w$ & 100 & 20 & 60 & 80 & 70 & 30 & 5 & 70 & 50\\
    & $d$ & 80 & 5 & 80 & 200 & 20 & 10 & 2 & 80 & 200\\
    & $k$ & 5 & 10 & 3 & 5 & 5 & 10 & 2 & 5 & 10\\
    \bottomrule
\end{tabular}
\label{tab:synthetic_parameters}
\end{table}

Table~\ref{tab:syn_datasets_characteristics} summarizes the characteristics of all the datasets used in our experiments, highlighting the number of samples contained in each dataset, and the number and shapes of the clusters in which those samples are grouped.

\begin{table}[t]
\caption{Best parameters found for the synthetic datasets. Column $P$ refers to the parameter names, while the numbered columns represent the datasets as follows: $10 = shapes$, $11 = zelnik1$, $12 = zelnik3$, $13 = zelnik6$, $14 = r15$, $15 = rings$, $16 = 3mc$, $17 = compound$, $18 = impossible$.}
\centering
\small
\begin{tabular}{l c r r r r r r r r r}
    Algorithm & P & 10 & 11 & 12 & 13 & 14 & 15 & 16 & 17 & 18
    \\\midrule
    \multirow{4}{*}{\dbstream} 
    & $r$ & 1 & .77 & .73 & .57 & .42 & .99 & .50 & 1.19 & .09\\
    & $C_m$ & 2 & 1.93 & 2.46 & 1.51 & 2.84 & 2.79 & 2.89 & 2.30 & 2.89\\
    & $\alpha$ & .1 & .07 & .13 & .16 & .89 & .67 & .04 & .31 & .01\\
    & $t_{gap}$ & 500 & 3.6k & 226 & 2.8k & 2k & 5k & 757 & 4.5k & 1.8k\\
    \midrule
    \multirow{4}{*}{\dstream}
    & $g_s$ & .30 & .13 & .10 & .13 & .21 & .15 & .11 & .07 & .19 \\
    & $t_{gap}$ & 1000 & 4.4k & 1.7k & 4.9k & 2.7k & 1.1k & 810 & 700 & 644 \\
    & $C_m$ & 3.00 & 1.63 & 2.43 & 1.24 & 1.62 & 2.37 & 2.71 & 2.29 & 1.34 \\
    & $C_l$ & .80 & .95 & .50 & 1.21 & 1.33 & 1.13 & 1.22 & .81 & .42 \\
    \midrule
    \multirow{2}{*}{\cedas}
    & $r$ & .2 & .1 & .15 & .1 & .02 & .01 & .1 & .01 & .01\\
    & $d$ & 20 &  2 &   2 &  8 &.  4 &   2 &  5 &   2 &   2\\
    \midrule
    \multirow{3}{*}{\compass}
    & $t_w$ & 50 & 5 & 5 & 5 & 20 & 80 & 70 & 80 & 50 \\
    & $d$ & 50 & 2 & 2 & 2 & 10 & 60 & 10 & 40 & 30 \\
    & $k$ & 1 & 3 & 3 & 3 & 2 & 10 & 5 & 1 & 3 \\
    \bottomrule
\end{tabular}
\label{tab:synthetic_parameters2}
\end{table}

Since the synthetic datasets do not represent data streams, during the experiments we consider data samples one by one to update the clustering models.
Moreover, given the time-related assumption (i.e., data samples that appear close in time refer to the same activity), we sort the data samples according to their class label.
Finally, Tables~\ref{tab:synthetic_parameters} and~\ref{tab:synthetic_parameters2} show the best parameter configuration for each algorithm found by using the grid-search technique.

\subsubsection{Results}

\begin{figure}[t]
    \centering
    \begin{subfigure}{.2\linewidth}
        \centering
        \begin{tikzpicture}
            \node[inner sep=0pt] (russell) at (0,0)
    {\includegraphics[width=\textwidth]{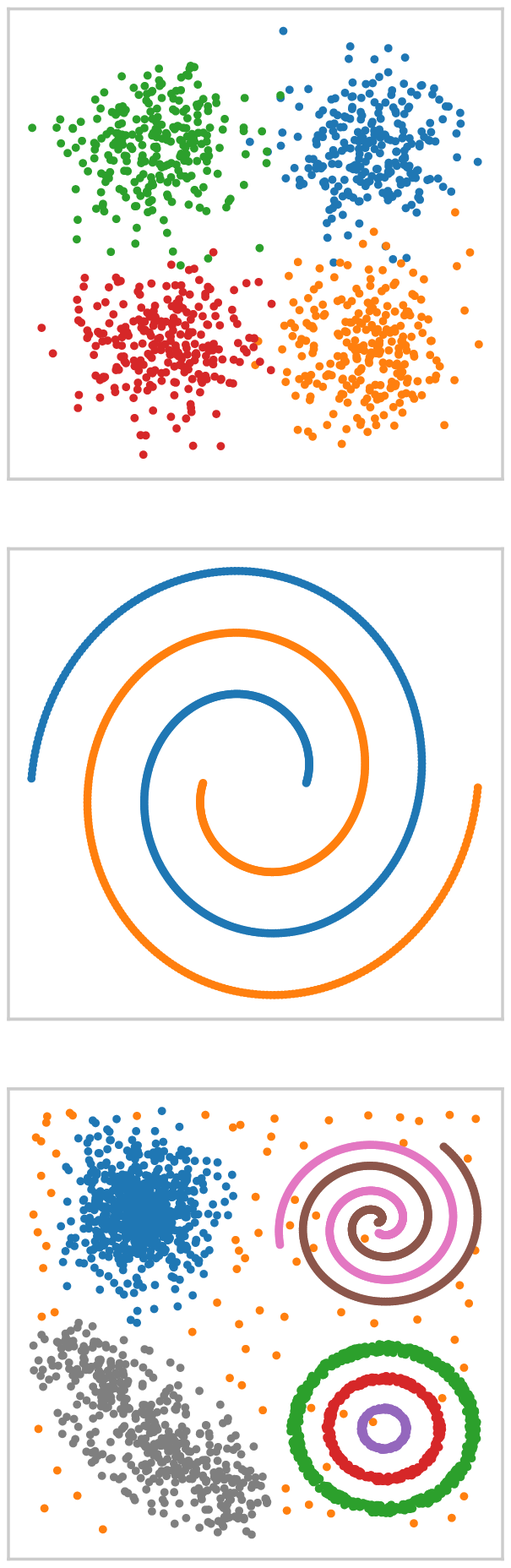}};
        \node[text width=1cm, rotate=90] at (-1.3,2) {square};
        \node[text width=1cm, rotate=90] at (-1.3,0) {spiral};
        \node[text width=1cm, rotate=90] at (-1.3,-2.4) {impossible};
        \end{tikzpicture}
        \caption{}
        \label{fig:syn_original}
    \end{subfigure}
    \hspace{0.15em}%
    \begin{subfigure}{.2\linewidth}
        \centering
        \includegraphics[width=\linewidth]{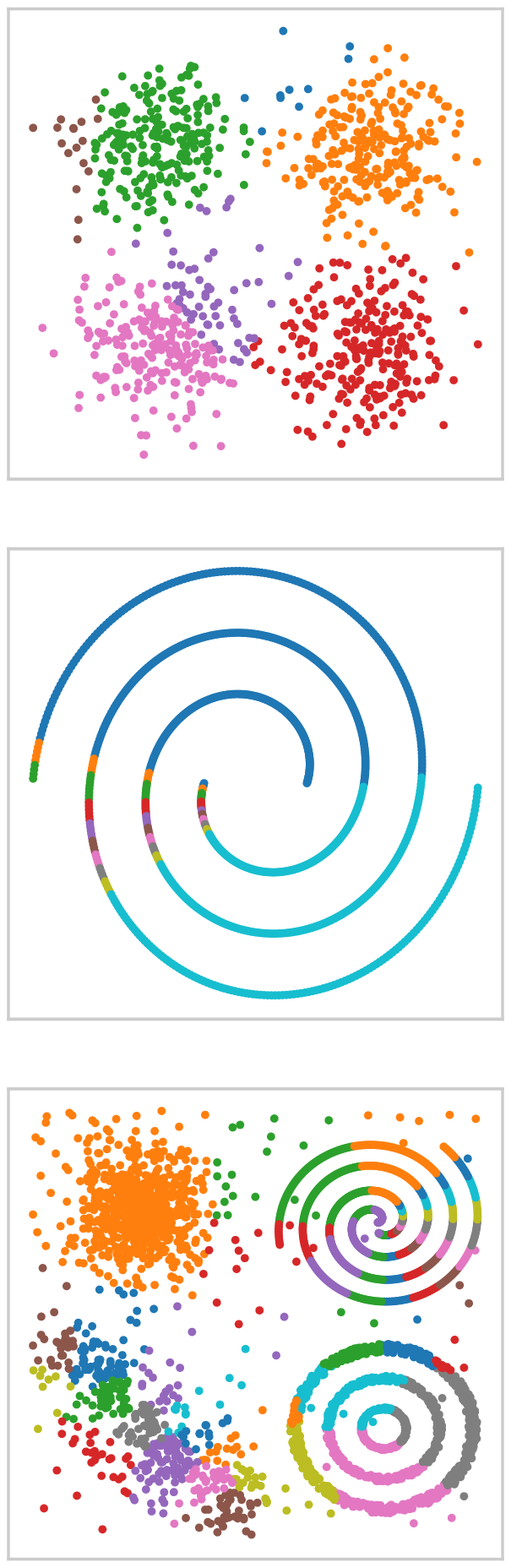}
        \caption{}
        \label{fig:syn_DBStream}
    \end{subfigure}
    \hspace{-0.5em}%
    \begin{subfigure}{.2\linewidth}
        \centering
        \includegraphics[width=\linewidth]{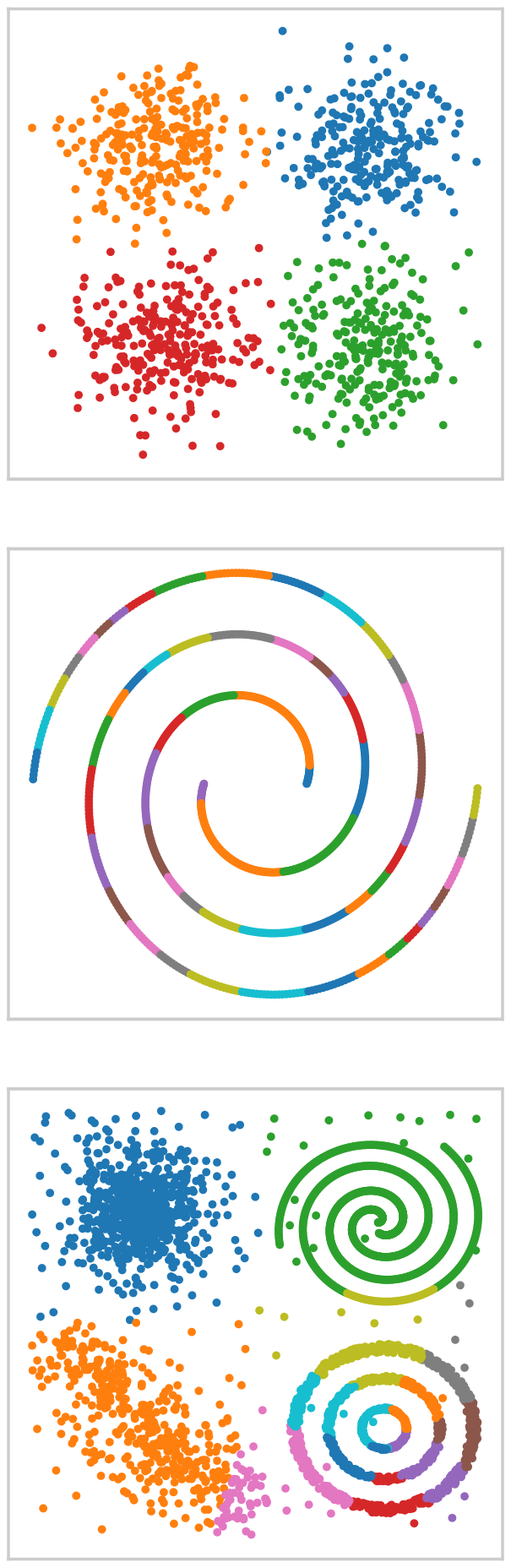}
        \caption{}
        \label{fig:syn_DStream}
    \end{subfigure}
    \hspace{-0.5em}%
    \begin{subfigure}{.2\linewidth}
        \centering
        \includegraphics[width=\linewidth]{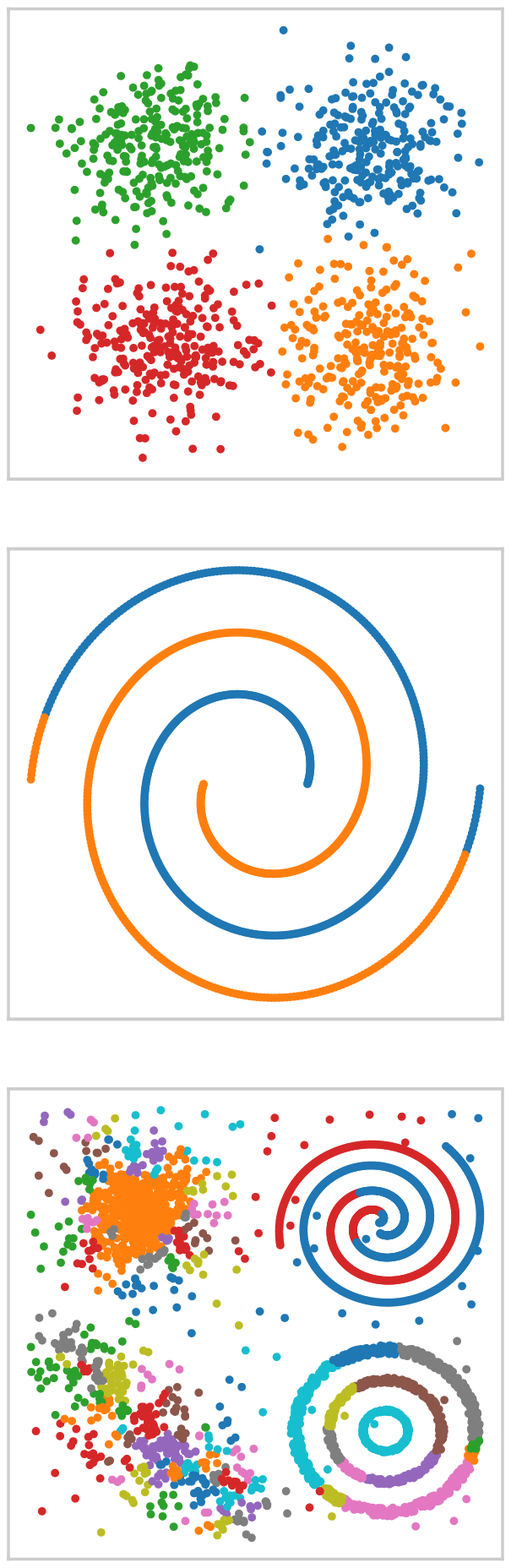}
        \caption{}
        \label{fig:syn_CEDAS}
    \end{subfigure}
    \hspace{-0.5em}%
    \begin{subfigure}{.2\linewidth}
        \centering
        \includegraphics[width=\linewidth]{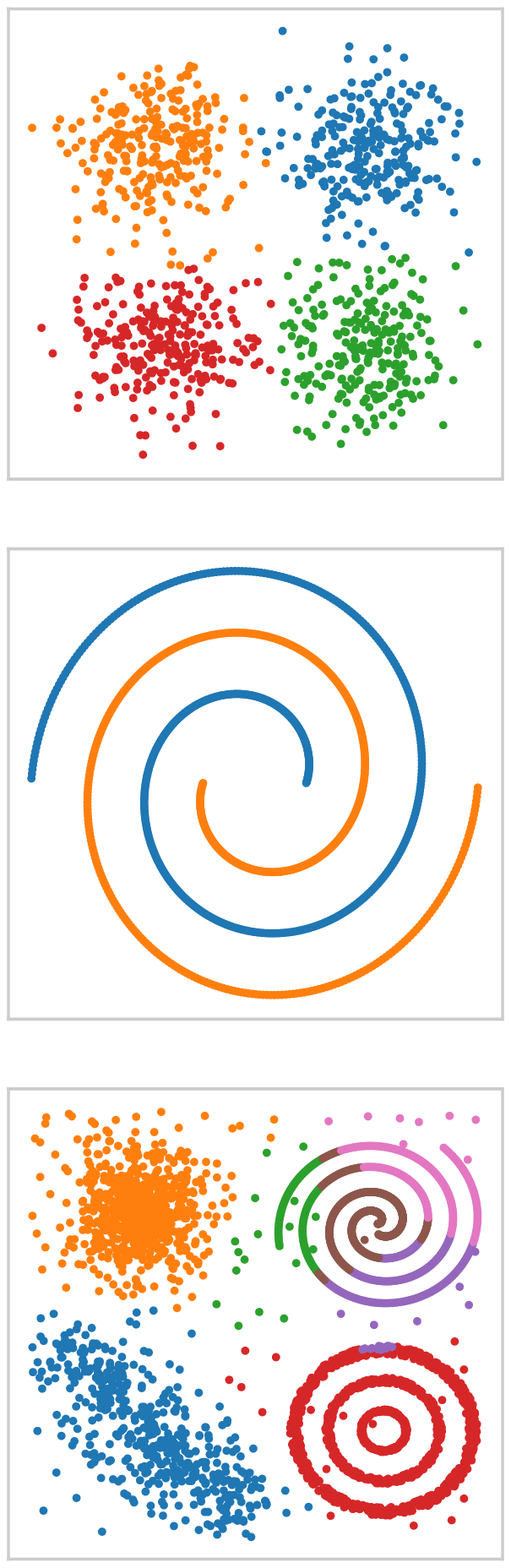}
        \caption{}
        \label{fig:syn_COMPASS}
    \end{subfigure}
    \caption{Representations of the clusters in the original datasets (a), and the clusters identified by \dbstream~(b), \dstream~(c), \cedas~(d), and \compass~(e). Best viewed in color.}
    \label{fig:syn_plots}
\end{figure}

Since the synthetic datasets contain bi-dimensional data points, we can easily plot the clusters configurations generated by the reference algorithms and graphically compare them with the ground truth.
Therefore, we start the discussion of the obtained results by analyzing the cluster configurations shown in Figure~\ref{fig:syn_plots}.
More specifically, Figures~\ref{fig:syn_DBStream}, \ref{fig:syn_DStream}, \ref{fig:syn_CEDAS}, and~\ref{fig:syn_COMPASS} show the graphical representation of the clusters created by \dbstream, \dstream, \cedas, and \compass~in 3 different datasets: square, spiral, and impossible.
As we can note, \compass~identifies the perfect configuration for both spherical and spiral clusters (i.e., \emph{square} and \emph{spiral} datasets), while the \dbstream~and \dstream~are not able to reconstruct the original cluster shapes, producing extremely fragmented results composed by several small clusters.
Even though \cedas~recognizes the spiral clusters better than both \dbstream~and \dstream, it does not reach the performances obtained by \compass.

In the experiment with the \emph{impossible} dataset, \compass~identifies the squared and rectangular clusters better than the others, but it fails in recognizing the spiral and concentric ones.
However, this result is not related to its ability to recognize those shapes, but it mainly depends on the sample order in the data stream.
Let us consider for example the concentric ring clusters in the \emph{impossible} dataset.
If the data points of the outermost circle are randomly distributed in the data stream (i.e., without following the cluster shape), points located on different sides of the circle may sequentially appear within the same time window.
For this reason, according to the time-based assumption, the algorithm will group those points, combining the outermost circle with the two inner ones and forming one single cluster.
Fortunately, it seems unlikely that such a cluster configuration may occur for human activities.
In the worst case, concentric groups with points very close to each other may describe different configurations of the same user's activity; therefore grouping them in the same cluster would be the most correct choice that the algorithm can take.

\begin{figure}[t]
    \begin{subfigure}{\textwidth}
        \centering
        \includegraphics[width=\textwidth]{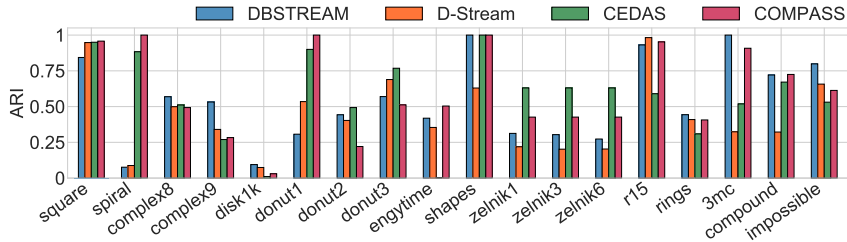}
        \caption{ARI}
        \label{fig:synthetic_results_ari}
    \end{subfigure}\\
    \begin{subfigure}{\textwidth}
        \centering
        \includegraphics[width=\textwidth]{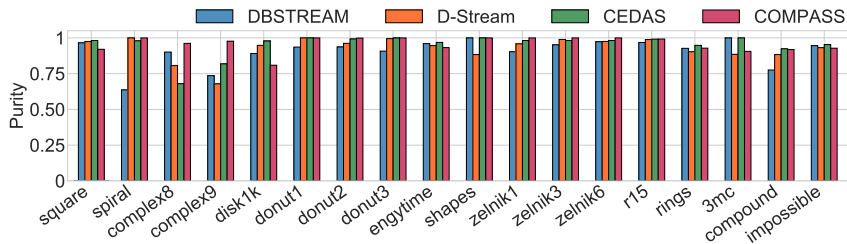}
        \caption{Purity}
        \label{fig:synthetic_results_purity}
    \end{subfigure}
    \caption{Synthetic datasets results.}
\end{figure}

In order to evaluate the performances of the three algorithms with more details, we compare their results in terms of both ARI of the Macro-clusters (Figure~\ref{fig:synthetic_results_ari}) and Purity of the micro-clusters (Figure~~\ref{fig:synthetic_results_purity}) created during the experiments.
As we can note, the obtained results show considerable differences among the different datasets.
According to~\citep{Carnein:2017:ECS:3075564.3078887}, this is due to the fact that the performances of online clustering algorithms strongly depend on the characteristics of the specific dataset.
In terms of ARI, both \compass~and \cedas~obtain the higher scores in 5 different experiments, \dbstream~achieves the best results with 6 datasets, while \dstream~performs poorly, obtaining the higher ARI scores in just one experiment (i.e., \emph{r15}).
In addition, in those experiments where the benchmark algorithms perform better than our solution, their ARI gain is not higher than 12\% (for both \dbstream~and \dstream) and 19\% (for \cedas).
On the contrary, when \compass~performs better than the reference algorithms it obtains, on average, results 15\% higher than \cedas, 25\% higher than \dbstream, and 35\% higher than \dstream.

Regarding the Purity results, all the algorithms are able to create micro-clusters with very high purity scores.
In this case, \compass obtains the best results in the majority of experiments (i.e., with 7 datasets), while \cedas~performs better in the remaining 6 experiments.
Although neither \dstream~nor \dbstream~perform better than the others, they achieve the same results as the best algorithm in 5 experiments: \dstream~with \emph{spiral}, \emph{donut1}, and \emph{donut3}, while \dbstream~with \emph{shapes} and \emph{3mc}.

These set of experiments show that \compass~is able to correctly identify clusters of different shapes and distributions, obtaining comparable or even better results than the state-of-the-art online clustering algorithms.
At this point, we can test \compass~in our reference scenario, evaluating its capability in identifying user complex activities from mobile sensors data streams.

\subsection{Real-world context datasets}

The main goal of \compass~is to find similarities in the mobile sensors data to recognize the situation in which a mobile user is involved.
In order to evaluate its performance in the reference scenario, in this section we perform a set of experiments by using two real-world datasets: \emph{ContextLabeler}, presented in our previous work~\citep{Campana:2018:LMU:3267305.3274178}, and \emph{ExtraSensory}, proposed in~\citep{Vaizman:2018:EAD:3173574.3174128}.
Both datasets have been collected from real mobile devices and in the same experimental setup: data has been collected from users that were engaged in their regular natural behavior (i.e., in the wild).
The differences between the two datasets mainly reside in the number of users involved in the sensing experiment and the type of the collected sensors features.
ExtraSensory has been created through a large scale experiment involving 60 users, but it mainly focuses on data related to physical sensors (e.g., accelerometer, GPS, and gyroscope) and device status (e.g., audio level and battery state).
On the other hand, ContextLabeler has been collected from just 3 participants, but it contains a broader set of information, including physical sensors and data characterizing both the interaction between the user and her device, and the surrounding environment.
Therefore, to get reliable results for our reference scenario, we perform two different sets of experiments by using both datasets.

Compared with the experiments performed with the synthetic datasets, in this case, we use a different approach to evaluate \compass~performances.
Specifically, we employ the \emph{horizon-based prequential evaluation} approach, which is a standard technique to evaluate online clustering algorithms~\citep{Aggarwal:2003:FCE:1315451.1315460, Carnein:2017:ECS:3075564.3078887}.
According to this approach, at a given evaluation time $t$, we test the current model with the next $H$ (horizon) samples in the data stream.
In a mobile scenario it is reasonable to assume that we are interested in monitoring the user's context at most every minute.
Therefore, in all the following experiments, we set $t=1$.
Note that the choice of $H$ does not affect the clustering results, but only the evaluation~\citep{Carnein:2017:ECS:3075564.3078887}.
In other words, by varying the horizon length, we can assess the algorithm performances in predicting the user's context among different time windows.
Therefore, since in both datasets a data sample represents a 1-minute snapshot of the current user context, we set $H = 20$, which allows us to evaluate the ability of the system in predicting the user's behavior in the next 20 minutes.
Considering the characteristics of the mobile environment, where the user's context might be extremely variable, setting $H=20$ enables us to assess the capability of COMPASS in self-adapting its model according to the changes in the user situation in our reference scenario.

In the next sections, we describe the main characteristics of the two datasets, the data preparation, the performed experiments and, finally, we discuss the obtained results.

\subsubsection{ContextLabeler Dataset}

\begin{figure}[t]
\centering
    \includegraphics[width=.99\linewidth]{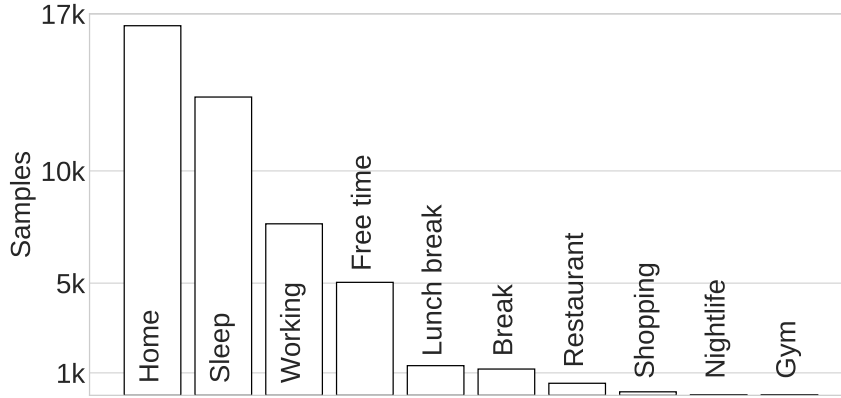}
\caption{Labels distribution.}
\label{fig:contextlabeler_labels}
\end{figure}

The \emph{ContextLabeler} dataset contains a total of 45681 data samples labeled by 3 voluntary users.
Each sample is characterized by 1332 features that represent a heterogeneous set of both physical and virtual sensors, which describe the status of the mobile device, the interactions between the user and her device, the surrounding environment, and other devices in proximity.

During a period of two weeks, the volunteers used a specific application installed on their smartphones that allowed them to freely specify their daily activities, while it was continuously collecting sensors data in the background.
Performing sensing experiments unobtrusively is crucial to effectively simulate a realistic mobile scenario.
In fact, according to~\citep{8334779}, most of the research studies in the area of activity recognition and human behavior modeling base their results on experiments performed in controlled environments (e.g., a research laboratory).
During the data collection process (often performed with the same device), volunteers are asked to perform some activities that have been previously defined by researchers.
However, in the real world, we have heterogeneous devices and different users may have different ways of doing the same activity; thus the experimental results usually diverge from those obtained in the lab~\citep{Mafrur2015}.
Therefore, to evaluate the proposed algorithm on data collected into the wild makes it more reliable for real-world applications.

Figure~\ref{fig:contextlabeler_labels} shows the distribution of the labels specified by the users.
It is worth noting that the distribution is highly skewed, where some labels are associated with a high number of data samples (e.g., \emph{Home} and \emph{Sleep}), while others (e.g., \emph{Nightlife} and \emph{Gym}) appear less frequently in the entire dataset.
This characteristic appears in many real-world datasets and it is usually known as ``unbalanced dataset problem'' because it can negatively influence the generalization and reliability of
supervised learning algorithms.
In fact, if we train a classifier by using a very skewed dataset, the learned model will be biased towards the majority class, which means that new observations will have more chances to be misclassified.
However, in this work, we are not interested in evaluating our proposal in terms of prediction accuracy, but we compare the clustering configuration generated by processing the data stream with the true class labels contained in the datasets.
Therefore, in this case, the skewed labels distribution does not affect the reliability of our experiments.

\subsubsection{ExtraSensory Dataset}
\label{sec:extrasensory}

\begin{figure}[t]
    \begin{subfigure}{.49\textwidth}
        \centering
        \includegraphics[width=\linewidth]{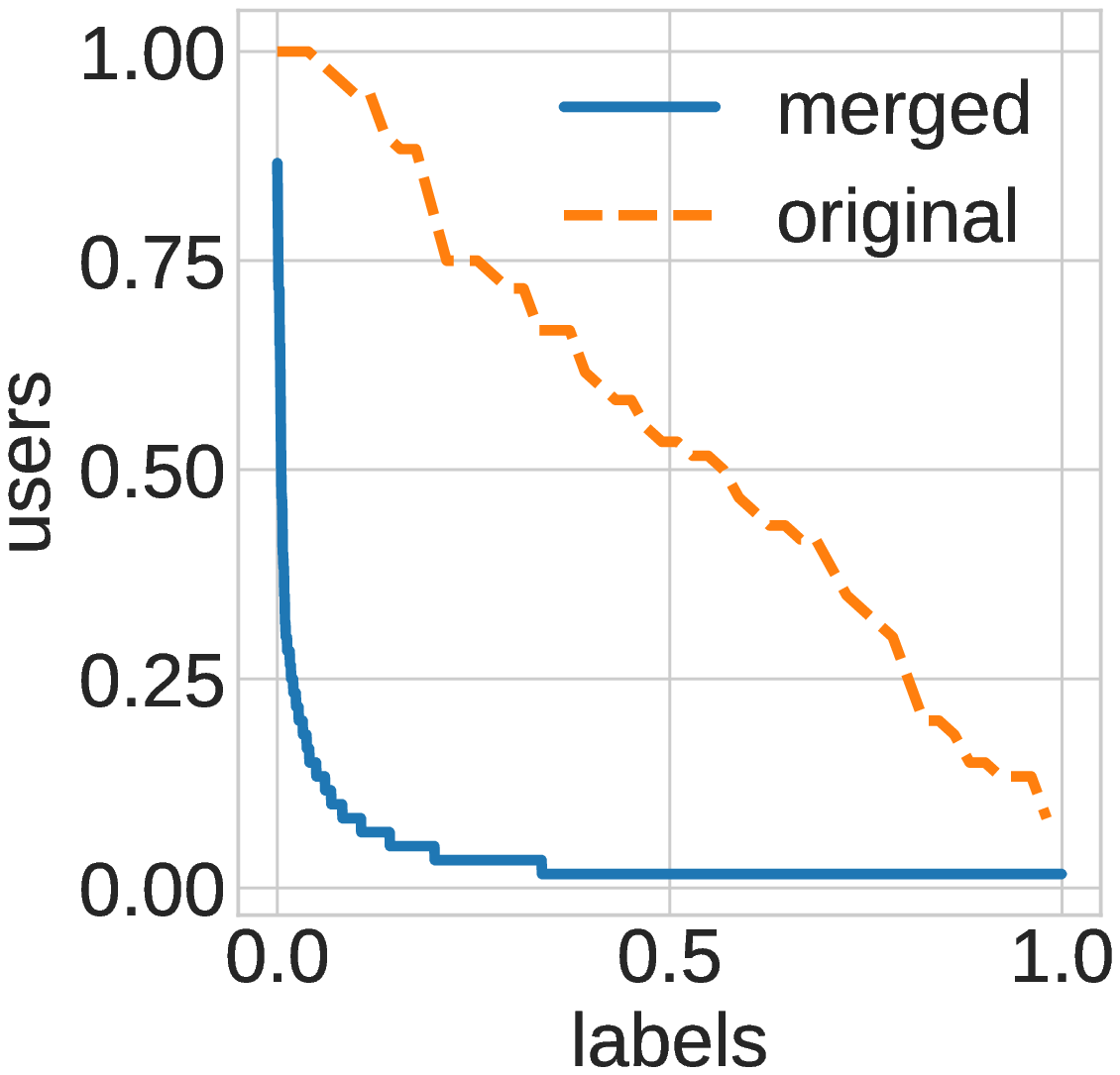}
        \caption{Labels distributions.}
        \label{fig:extasensory_label_distribution}
    \end{subfigure}
    \begin{subfigure}{.49\textwidth}
        \centering
        \includegraphics[width=\linewidth]{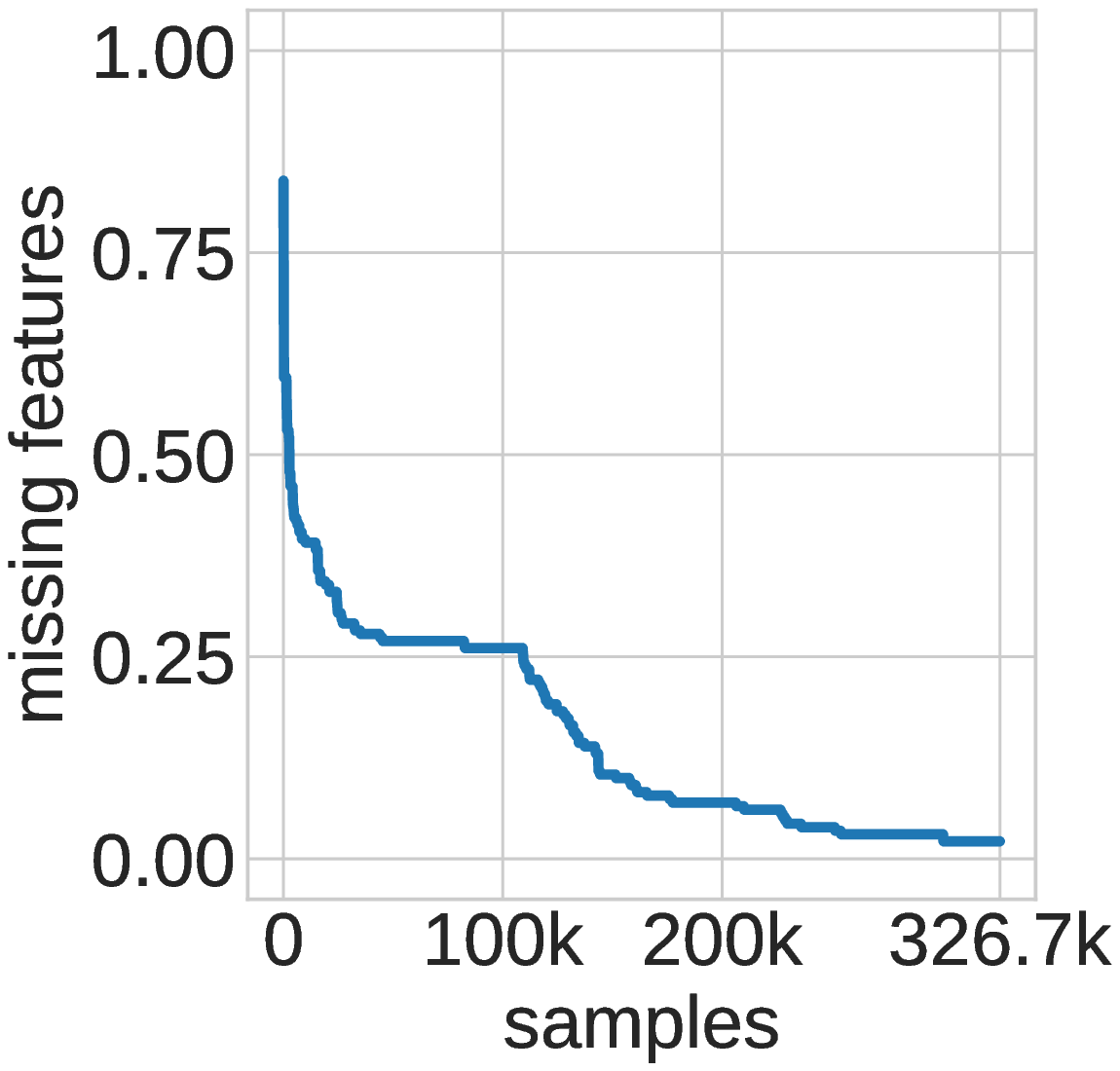}
        \caption{Missing data.}
        \label{fig:extrasensory_missing_data}
    \end{subfigure}
    \caption{ExtraSensory dataset statistics.}
\end{figure}

The ExtraSensory dataset is similar to ContextLabeler, but it contains over 300 minutes of labeled sensing data generated by a wider set of users.
It is composed by 326687 data samples collected from 60 participants in their regular environments by using their own mobile devices and, in some cases, wearable devices (i.e., Pebble watch~\footnote{Smartwatch produced by Pebble Technology Corporation. Nowadays, its development is discontinued.}) to capture additional user-device interactions and to increase the sensing capabilities of the entire experiment.
Each data sample is composed by 228 features characterizing different physical and virtual sensors, including accelerometer, gyroscope, user location, and phone state (e.g., app status, battery state, and Wi-Fi availability).
Besides, during the sensing experiment, users could specify two types of labels to describe their current behavior: (i) a \emph{main activity} label among 7 possible choices describing movement and posture of the user (e.g.,  \emph{lying down}, \emph{sitting}, or \emph{standing in place}); and (ii) one or more \emph{secondary activity} labels that describe more specific context in different aspects, i.e., sports (e.g. \emph{playing basketball}, \emph{at the gym}), transportation (e.g. \emph{drive - I'm the driver}, \emph{on the bus}), basic needs (e.g. \emph{sleeping}, \emph{eating}, \emph{toilet}), social environment (e.g. \emph{with family}, \emph{with co-workers}), location (e.g. \emph{at home}, \emph{at work}, \emph{outside}), etc.
The total number of unique labels contained in the dataset is 51.

Since we are not interested in multi-label prediction or classification, for each data sample we combine all the labels specified by the user in one single label.
For example, if a given data sample is associated with the labels \emph{SITTING}, \emph{PHONE\_IN\_POCKET} and \emph{AT\_SCHOOL}, we create the single label \emph{SITTING\_PHONE\_IN\_POCKET\_AT\_SCHOOL}, that describes the general situation in which the user was involved when her device collected the corresponding sensors data.
After the combination of the original labels, the total number of unique labels in the dataset increased up to 2049.
Figure~\ref{fig:extasensory_label_distribution} shows the labels distribution in terms of the percentage of participants who have used the labels to specify their current context before (\emph{original}) and after (\emph{merged}) the pre-processing step.
It is worth noting that this process significantly changes the labels distribution and generally decreases the percentage of users who have used a specific label.
This is due to the fact that, by merging the labels associated with the same data sample, we specify the context described by the sensors data with more details.
Indeed, the merged labels are characterized by a long-tailed distribution, where only a small amount of specific situations (i.e., 3\% of the labels) have been shared by several users (i.e., greater than 25\% of the number of users), while the majority of the labels have been used only by a small number of participants.

Unfortunately, the ExtraSensory dataset is also characterized by the presence of a considerable amount of missing data.
This is mainly because, during the data collection, participants were able to turn off specific sensors to save battery of their mobile devices.
Figure~\ref{fig:extrasensory_missing_data} shows the percentage of missing values in each data sample.
According to the plot, none of the data samples in the dataset is complete (i.e., without missing values).
For this reason, removing samples with missing values is not a feasible option.
Therefore, as a final preprocessing step before testing our clustering algorithm against the three reference solutions, we impute missing data by using the Multiple Imputation technique, one of the most common approaches to tackle this problem~\citep{doi:10.1177/0962280216666564}.
According to this technique, we model each feature with missing values as a function of other features, and we use that estimate for iteratively imputing missing data.
More specifically, at each iteration, a feature is designated as output \emph{y} and the others are treated as input \emph{X}.
Then, a regressor is fitted on the set of samples $(X, y)$ and, finally, it is used to predict the missing values of $y$, thus obtaining a suitable dataset for our experiments.

\subsubsection{Results}

\begin{table}[t]
\caption{Best parameters found for the real-world datasets.}
\small
\centering
\begin{tabular}{ l c r r}
    Algorithm & Parameter & ContextLabeler & ExtraSensory
    \\\midrule
    \multirow{4}{*}{\dbstream} 
    & $r$       & 0.66   & 0.03  \\
    & $C_m$     & 2.27   & 2.87  \\
    & $\alpha$  & 0.30   & 0.35  \\
    & $t_{gap}$ & 3.1k   & 3.9k  \\
    \midrule
    \multirow{4}{*}{\dstream}
    & $g_s$     & 0.27   & 0.14  \\
    & $t_{gap}$ & 4.9k   & 3.1k  \\
    & $C_m$     & 1.93   & 1.27  \\
    & $C_l$     & 0.04   & 0.14  \\
    \midrule
    \multirow{2}{*}{\cedas}
    & $r$       & 0.4    & 0.8    \\
    & $d$       & 5      & 5      \\
    \midrule
    \multirow{3}{*}{\compass}
    & $t_w$     & 600    & 600 \\
    & $d$       & 5      & 5     \\
    & $k$       & 2      & 5     \\
    \bottomrule
\end{tabular}
\label{tab:real_parameters}
\end{table}
\begin{figure*}[t]
    \begin{subfigure}{.9\textwidth}
        \centering
        \includegraphics[width=\linewidth]{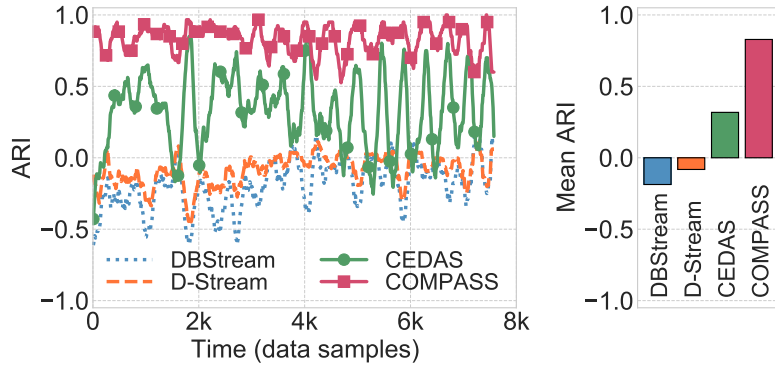}
        \caption{ContextLabeler ARI}
        \label{fig:contextlabeler_ari}
    \end{subfigure}\\
    \begin{subfigure}{.9\textwidth}
        \centering
        \includegraphics[width=\linewidth]{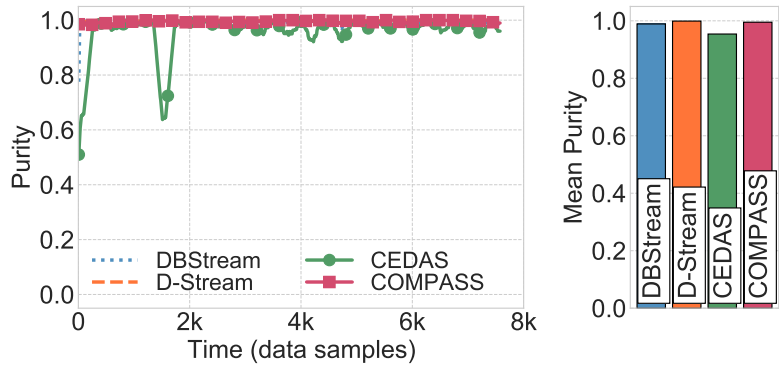}
        \caption{ContextLabeler Purity}
        \label{fig:contextlabeler_purity}
    \end{subfigure}
    \caption{Results obtained with the ContextLabeler dataset.}
    \label{fig:cl_results}
\end{figure*}
\begin{figure*}[t]
    \begin{subfigure}{.9\textwidth}
        \centering
        \includegraphics[width=\linewidth]{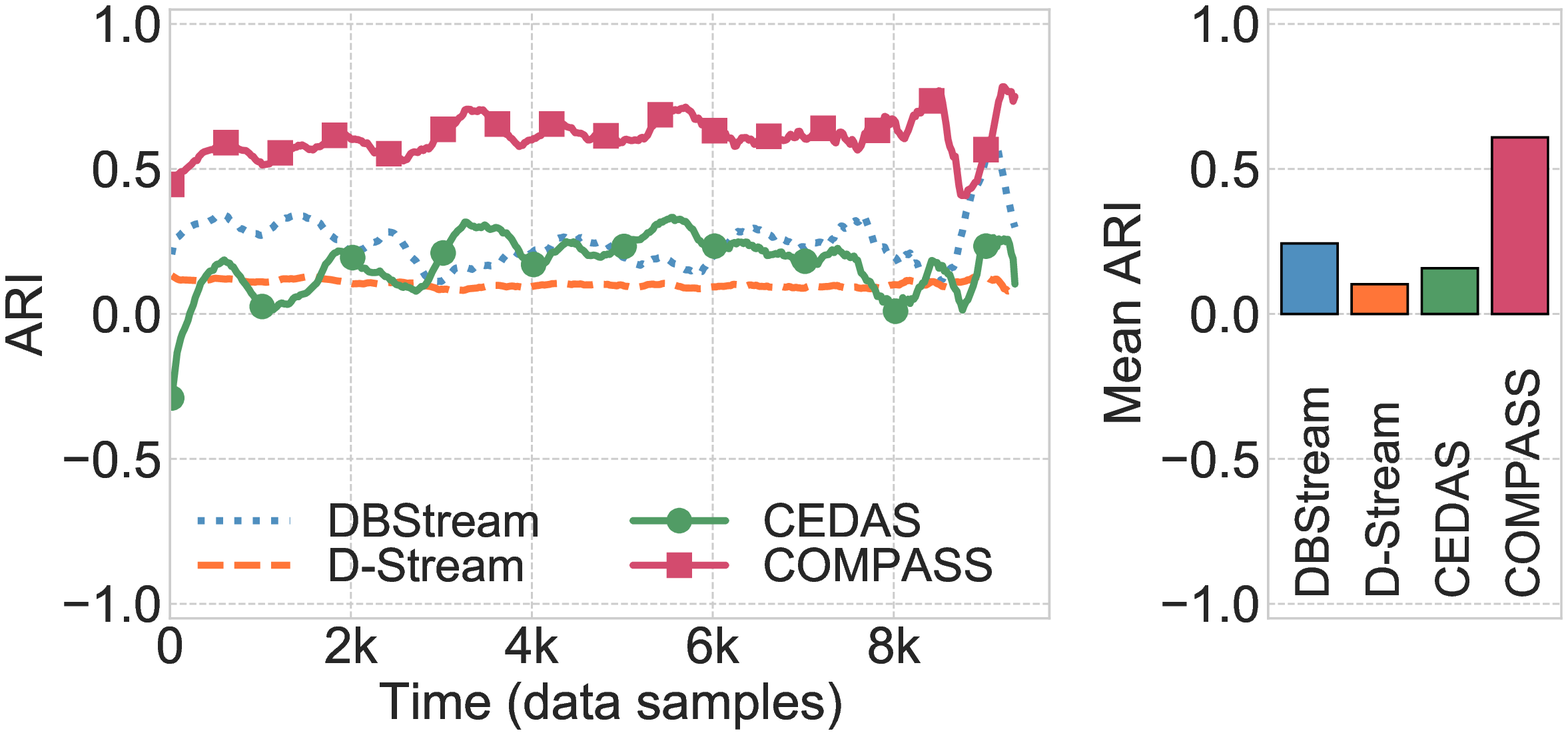}
        \caption{ExtraSensory ARI}
        \label{fig:extrasensory_ari}
    \end{subfigure}\\
    \begin{subfigure}{.9\textwidth}
        \centering
        \includegraphics[width=\linewidth]{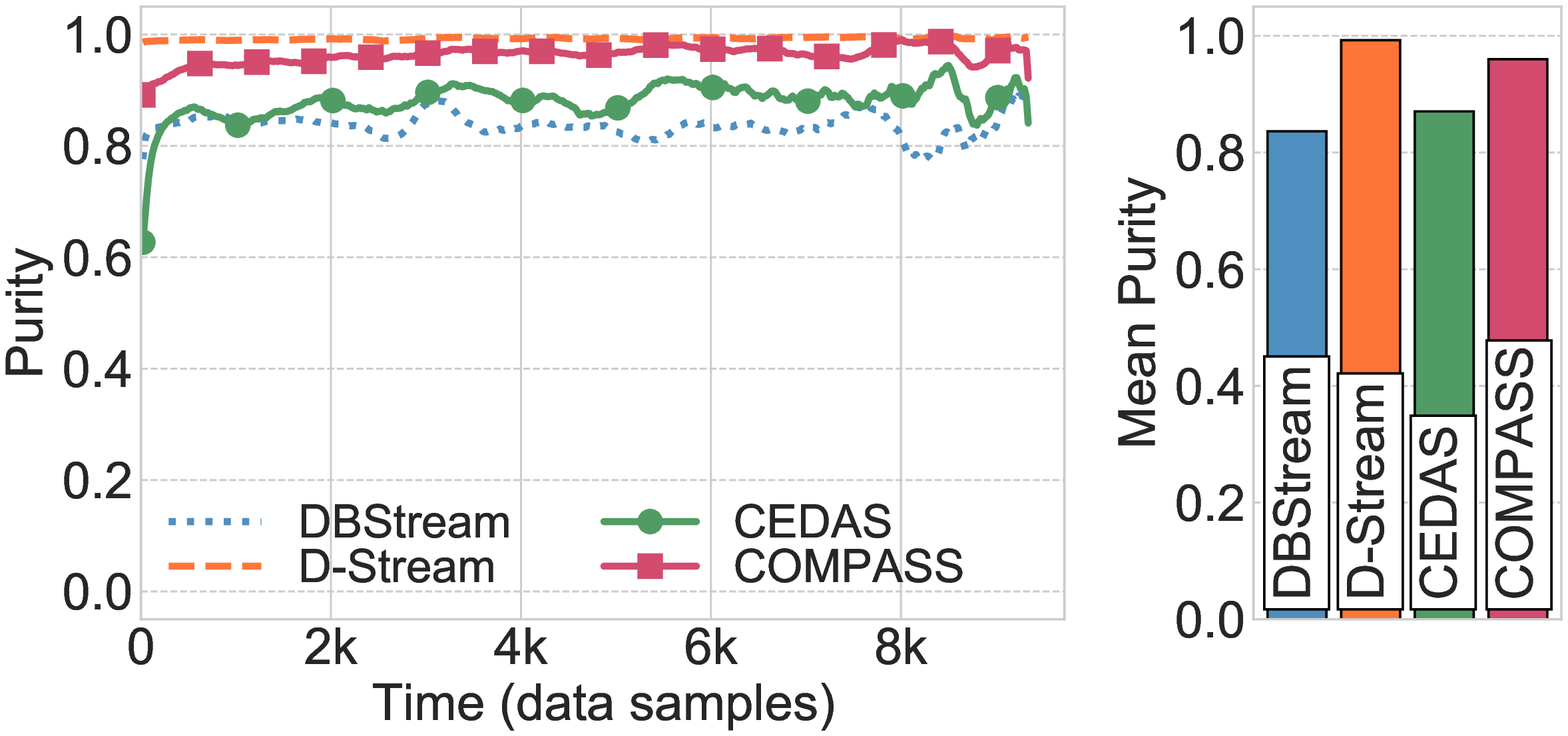}
        \caption{ExtraSensory Purity}
        \label{fig:extrasensory_purity}
    \end{subfigure}
    \caption{Results obtained with the ExtraSensory dataset.}
    \label{fig:es_results}
\end{figure*}

Different users may perform the same activity in different ways.
In order to build one single model (i.e., a \emph{global model}) that can achieve good results for all the users, we perform the following steps.
First, we estimate the best model parameters for each user by using the grid search approach.
Then, we select the average value of the best parameters to create a global model. 
Finally, we evaluate the performance of the global model in terms of both ARI and Purity by performing the prequential evaluation task.
Table~\ref{tab:real_parameters} shows the parameters of the global models used in both ContextLabeler and ExtraSensory experiments.

Figures~\ref{fig:cl_results} and~\ref{fig:es_results} show the results in terms of both ARI of the Macro-clusters and Purity of the micro-clusters generated by the three algorithms during the online evaluation with the two datasets.
More specifically, the figures show the running average results obtained by the global models over all the users contained in the datasets and the average results over the entire stream.
\compass~obtains the best results in terms of both the reference evaluation metrics.

In the experiments performed with the ContextLabeler dataset (Pictures~\ref{fig:contextlabeler_ari} and~\ref{fig:contextlabeler_purity}), all the algorithms obtain a mean Purity score close to $0.99$, except for \cedas~that reaches only $0.95$.
However, according to the results shown in Picture~\ref{fig:contextlabeler_ari}, both \dstream~and \dbstream~definitely fail in identifying the correct clusters configuration.
In fact, they obtained an average ARI score of $-0.19$ and $-0.08$, respectively.
According to the ARI definition~\citep{hubert1985comparing}, the obtained results mean that, compared with the ground truth, the clustering configurations found by both \dstream~and \dbstream~are comparable with the results obtained by a random guesser.
\cedas~performs better than the previous algorithms, obtaining a mean ARI of 0.32.
However, the use of a fixed radius highlights the main limitations of the algorithm.
Indeed, the fluctuating results obtained during the entire simulation clearly indicate that the algorithm fails in adapting its behavior to the changes in the data stream.
On the contrary, \compass~outperforms the other solutions in terms of ARI, keeping the highest score during the entire data stream processing.
In fact, the mean ARI score obtained by our solution is $0.83$, which corresponds to an increase of $1.02$ ($125.93\%$) with respect to \dbstream, $0.91$ ($98.91\%$) of the \dstream~performance, and $0.52$ ($38.64\%$) of the mean results obtained by \cedas.

In the ExtraSensory experiments (Pictures~\ref{fig:extrasensory_ari} and~\ref{fig:extrasensory_purity}), \dstream~creates the purest micro-clusters by obtaining a mean Purity score of $0.96$, but it is not able to reconstruct the correct configurations of the original clusters in the stream (i.e., the mean ARI is close to zero, $0.1$).
On the other hand, \dbstream~creates less pure micro-clusters compared with \dstream~(i.e., it scores $0.84$ of mean Purity), but it creates a slightly better configuration of the Macro-clusters, by obtaining an average ARI of $0.24$.
Compared with the previous experiment, \cedas~results are more steady, but its performance is comparable with both \dbstream~and \dstream, obtaining a mean ARI of $0.16$ and a mean Purity of $0.87$.
Finally, also in this second set of experiments, \compass~outperforms the reference algorithms by obtaining an average ARI value of $0.61$.
In other words, our solution improves the performance of \dbstream, \dstream, and \cedas~by $0.37$ ($29.03\%$), $0.51$ ($45.45\%$), and $0.45$ ($37.03\%$), respectively.
\compass~performs particularly well also in terms of micro-clusters purity, by reaching a mean score very close to the best result (i.e., $0.96$), which results in a performance increase of $7.1\%$ against \dbstream~and $4.81\%$ compared with \cedas.

\subsection{Time performances evaluation}

Our main objective is to distinguish the user contexts by processing sensors data directly on the local device, relying only on its hardware capabilities and computational resources.
This is due to the fact that the mobile environment can be extremely dynamic, and the user context can change multiple times in just a few minutes.
Even though the identification of the user context can be performed on remote servers by using offloading mechanisms~\citep{8024034}, the delay of the data transmission may make the context identification useless with respect to the service optimization, since the user could have changed her context in the meanwhile.
In addition, the use of remote servers or cloud-based services may demotivate privacy-aware users to use context-aware services, since a third-party entity will be in charge of storing and managing users' personal data~\citep{6187862, ryan2011cloud}.
Therefore, the context-aware system should be executed on the local device and it should be able to provide the fastest response as possible.

\begin{figure}[t]
    \begin{subfigure}{.49\textwidth}
        \centering
        \includegraphics[width=\linewidth]{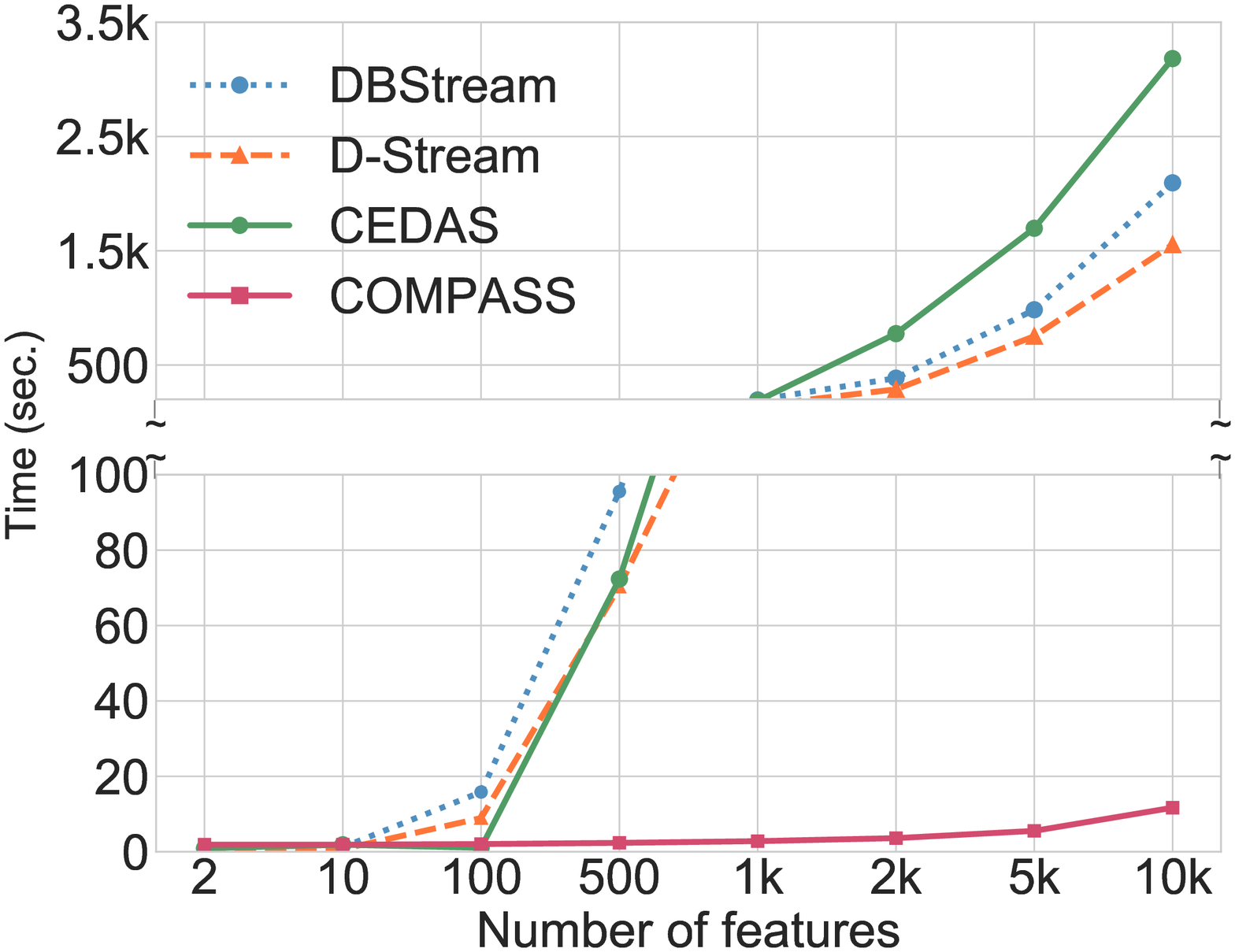}
        \caption{}
        \label{fig:complexity_all}
    \end{subfigure}
    \begin{subfigure}{.49\textwidth}
        \centering
        \includegraphics[width=\linewidth]{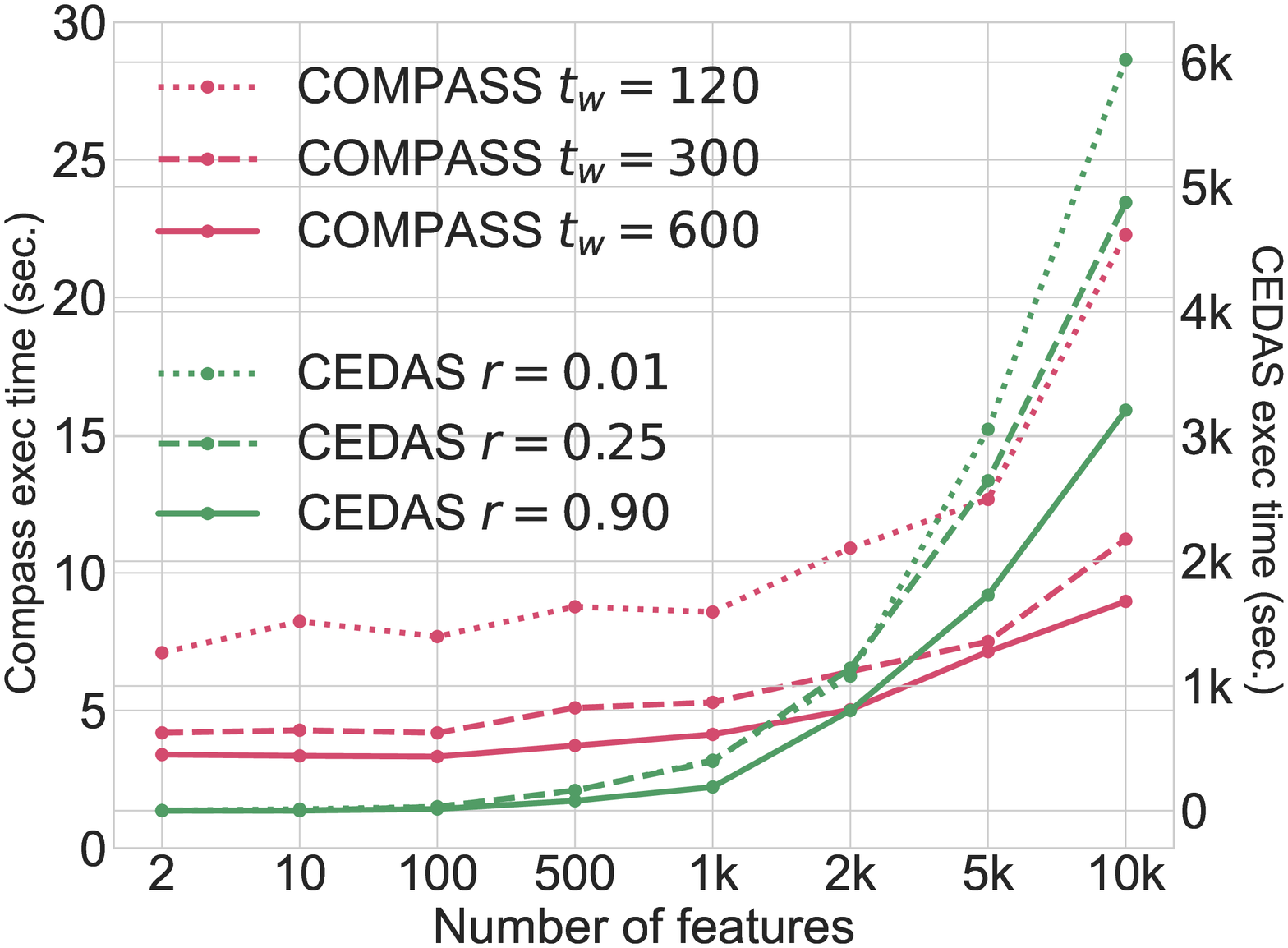}
        \caption{}
        \label{fig:complexity_cedas_compass}
    \end{subfigure}
    \caption{Time performances for different data dimensions: (a) shows the comparison among all the reference solutions by using the best parameter values found for the ExtraSensory dataset (see Table~\ref{tab:real_parameters}), while (b) compares \compass~and \cedas~with different configurations.}
\end{figure}

In this section, we perform an empirical evaluation of \compass~performances in terms of execution time and we compare the obtained results with the performances of our baseline algorithms: \dbstream~and \dstream.
To this aim, we have generated a set of 8 synthetic datasets by using the \texttt{dataset.make\_blobs} module provided by \emph{scikit-learn}, the popular machine learning library for Python~\footnote{\url{https://scikit-learn.org}}.
Since the user context can be represented by high-dimensional features vectors, we want to evaluate the time required by each algorithm to process a number of data samples characterized by a different number of features.
More precisely, each dataset is composed by 10000 data samples divided into 4 different Gaussian blobs that represent the clusters.
Moreover, each dataset contains samples characterized by a different number of features: 2, 10, 100, 500, 1000, 2000, 5000, and 10000 features.
We run the experiments on a GNU/Linux server with 64 CPUs and 132 GB of RAM.
However, in order to simulate a resource constrained environment, we run each algorithm-dataset configuration ten times on a single thread.

Figure~\ref{fig:complexity_all} shows the average time required by the clustering algorithms to process each of the 8 synthetic datasets by using the best parameters found in the last experiment (i.e., ExtraSensory).
It is worth noting that none of the four algorithms shows a linear complexity, but \compass~proved to be extremely faster, obtaining results with two orders of magnitude less than the baseline algorithms.
In fact, starting from data samples with 100 features, the reference solutions are characterized by particularly low performances in terms of execution times, requiring 
between 25 minutes (1553 seconds of \dstream) and 53 minutes (3181.99 seconds of \cedas) to process a dataset composed by data samples with 10000 dimensions, while \compass~requires less than 15 seconds.
The difference between the performances of \compass~and those obtained by \cedas~is certainly due to the use of a dynamic radius instead of a fixed one.
Indeed, using the same radius involves creating more micro-clusters, which slows down other procedures of the algorithm (e.g., finding the closest cluster for a new data sample and merging two or more clusters into a single Macro-cluster).
Figure~\ref{fig:complexity_cedas_compass} shows this characteristic in detail, reporting the execution times of both \cedas~and \compass~for various configurations: different sizes of the time window $t_w$ for CEDAS, and distinct radii $r$ for COMPASS.
As we can note, the choice of $r$ considerably affects the time performance of CEDAS, approximately requiring twice as long to process the entire dataset by using a small radius (i.e., $r=0.01$) instead of a large one (i.e., $r=0.9$).
Similarly, COMPASS performances depend on the size of $t_w$, but the gap between the worst and best configurations is 13.34 seconds, while for CEDAS is 46 minutes.

The obtained results clearly demonstrate the great advantage of the approach adopted by \compass~compared with the state-of-the-art.
Compared with \dbstream~and \dstream, our proposal identifies the Macro-clusters as soon as new data points appear in the stream, without running an offline phase to re-clustering the already identified micro-clusters.
Moreover, dynamically adapting its behavior to the characteristics of the new observations, compared with \cedas, \compass~reduces the model complexity, avoiding tuning the radius size and setting its value more naturally for our reference scenario.
In conclusion, this characteristics make \compass~the best candidate solution to provide context-aware features on mobile devices, processing the context data as soon as they appear in the sensors data stream.

% CONCLUSIONS ------------------------------------------------------------------------------------------------

\section{Conclusions and future work}
\label{sec:conclusions}

In this work, we presented \compass, a novel unsupervised and online algorithm to identify human complex activities, based on the stream of high-dimensional data generated by smartphone sensors.
More precisely, \compass~is an online clustering algorithm that identifies the changes in the user's context by exploiting the relation of time-dependency between data samples that describe the same user context, without requiring the collection of a large amount of annotated data for the training phase.
Moreover, it can extract an arbitrary number of clusters from the data stream without defining the number of expected classes in advance.
In this way, \compass~overcomes the main limitations of current context-recognition solutions, by identifying the user context as soon as mobile sensors generate new data.
These characteristics allow it to support a broad set of mobile applications, without requiring experts' knowledge.

We evaluated the effectiveness of the proposed solution performing 3 different sets of experiments by using a combination of both synthetic and real-world datasets.
First, we evaluated the ability of \compass~to recognize clusters of different shapes and densities by using 18 datasets, typically used to benchmark clustering algorithms.
Then, we employed two public datasets composed by sensors data and collected from a total of 63 smartphones to evaluate its performance in real mobile scenarios.
According to the obtained results, \compass~outperforms the state-of-the-art algorithms in online clustering in terms of both configuration and purity of the clusters identified from the data stream (i.e., ARI and Purity metrics).
Finally, we performed an empirical evaluation of its execution time.
We aim to bring context-aware features on devices with limited computational resources, thus the execution time is a crucial aspect that must be optimized.
Also in this case, \compass~proved to be extremely efficient and able to be executed on mobile devices.
In fact, it requires less than 15 seconds to evaluate 10000 data samples composed by 10000 features, while the reference solutions require more 30 min to process the entire dataset.

The definition of \compass~paves the way to additional research activities.
As a first step we are planning to test and evaluate the proposed solution in real mobile environments.
To this aim, we are currently integrating \compass~in ContextKit, a prototype mobile framework we developed to provide context-aware features to third-party mobile applications, presented in our previous work~\citep{Campana:2018:LMU:3267305.3274178}.
Currently, ContextKit implements a sensing layer that is able to collect data generated by a heterogeneous set of smartphone sensors, including both physical and virtual sensors.
We intend to extend ContextKit by creating a reasoning layer that implements \compass~to infer the user's context based on the data samples coming from the lower sensing layer.
Then, by including ContextKit in a mobile application, we will be able to evaluate the performances the entire framework in a real-world setting based also on the user's feedback.

As a second line of research, we are planning to investigate the possibility to improve the context recognition process by using fewer sensors data.
The most common techniques proposed in the literature to reduce the input space are \emph{feature selection}~\citep{CHANDRASHEKAR201416} and \emph{dimensionality reduction}~\citep{6918213}.
While the former aims at selecting the most relevant features, the latter finds a low-dimensionality representation that retains some meaningful properties of the original data.
Even though such methodologies are remarkably useful to optimize different aspects of machine learning algorithms (e.g., speeding-up the training times, simplifying the model, and enhancing its generalization), they are commonly used to reduce only the input feature space.
On the contrary, our goal is to use (or extend, if necessary) such techniques to select the most relevant mobile sensors that describe the user's context in mobile environments, thus performing a ``sensor selection'' task.
This will lead to two main advantages: (i) similarly to the standard techniques, it will be possible to improve the clustering model by reducing the data space, and (ii) to save the device resources by activating only those sensors that are necessary to characterize the user's context.

We are also studying additional mechanisms to further reduce the need for a tuning phase of the involved parameters.
In fact, searching for the best parameter values is a time-consuming task and, if the dataset used in the experiments is not very representative of the application scenario, it can lead to bad results once the solution is employed in real-world settings.
Even though \compass~is able to dynamically set one of the most important parameters of a clustering algorithm (i.e., the cluster radius size), other relevant parameters still remain to be set, for example, the micro-cluster shell and kernel areas sizes.
These two parameters control the creation of new Macro-clusters: if the shell area of the micro-cluster $a$ overlaps the kernel area of a micro-cluster $b$, \compass~assumes that they represent the same concept and then, the two are merged to create a single and bigger cluster.
To this aim, we will investigate new approaches to automatically set those parameters according to the incoming data samples, especially how they affect the performances of \compass~in finding the clustering setting that best represents the characteristic of the context data stream.

Finally, we will investigate the use of different similarity and distance metrics.
As highlighted in~\citep{7853264}, the choice of a proper metric is crucial in clustering analysis to reveal the natural grouping in a given dataset.
Specifically, as part of our future works, we are planning to explore the use of different metrics in \compass, evaluating their impact on the algorithm's ability to identify the user's contexts in the mobile environment.

% ACKS ------------------------------------------------------------------------------------------------

\section{Acknowledgments}
\label{sec:acks}

The authors would like to thank Matthias Carnein of the University of M{\"u}nster, Germany, for his useful suggestions and valuable support related to the evaluation protocol of stream clustering algorithms.
This work has been partially funded by the European Commission under H2020-INFRAIA-2019-1SoBigData-PlusPlus project. Grant number: 871042

% REFERENCES ------------------------------------------------------------------------------------------------

\section*{References}
\bibliography{paper.bib}

\end{document}